\pdfoutput=1

\documentclass[11pt]{article}

\usepackage{ACL2023}
 
\usepackage{times}
\usepackage{latexsym}
\usepackage{graphicx} 
\usepackage[T1]{fontenc}

\usepackage[utf8]{inputenc}

\usepackage{microtype}
\usepackage{subfigure}
\usepackage{inconsolata}
\usepackage{xspace}
\newcommand{\KANMT}{continual KD for NMT\xspace}

\usepackage{times}
\usepackage{latexsym}
\usepackage{makecell}
\usepackage{amsmath}
\usepackage{CJKutf8}
\usepackage{graphicx} 
\usepackage{threeparttable}
\usepackage{booktabs}
\usepackage{diagbox}
\usepackage{microtype}
\usepackage{multirow}
\usepackage{bbm}
\usepackage{color}
\usepackage{xspace}
\usepackage{dcolumn}
\usepackage{bm}
\newcolumntype{.}{D{.}{.}{-1}} 
\def\Bold #1.#2 {\multicolumn{1}{D{.}{\mathbf{.}}{-1}}{\mathbf{#1}.\mathbf{#2}}}
\def\BoldTwo #1.#2 {\multicolumn{1}{D{.}{\mathbf{.}}{-1}|}{\mathbf{#1}.\mathbf{#2}}}
\def\Underline #1.#2 {\multicolumn{1}{D{.}{\underline{.}}{-1}}{\underline{#1}.\underline{#2}}}
\def\It #1.#2 {\multicolumn{1}{D{.}{\mathit{.}}{-1}}{\mathit{#1}.\mathit{#2}}}
\usepackage{url}

\def\argmax{\mathop{\rm argmax}}%
\def\argmin{\mathop{\rm argmin}}%
\def\max{\mathop{\rm max}}%
\makeatletter
\renewcommand\@makefntext[1]{\leftskip=0em\hskip1em\@makefnmark#1}
\makeatother

\title{Continual Knowledge Distillation for Neural Machine Translation}

\author{
  Yuanchi Zhang$^{1,3}$, Peng Li$^{*2,4}$, Maosong Sun$^{1,3}$, Yang Liu\thanks{\:\: Corresponding authors.}$^{\:\:\:1,2,3,4}$   \\
  $^1$Dept. of Comp. Sci. \& Tech., Institute for AI, Tsinghua University, Beijing, China \\
  $^2$Institute for AI Industry Research (AIR), Tsinghua University, Beijing, China \\
  $^3$Beijing National Research Center for Information Science and Technology \\
  $^4$Shanghai Artificial Intelligence Laboratory, Shanghai, China \\
  \texttt{yuanchi-21@mails.tsinghua.edu.cn; lipeng@air.tsinghua.edu.cn} \\
  \texttt{\{sms,liuyang2011\}@tsinghua.edu.cn}
}

\begin{document}
\maketitle
\begin{abstract}
  While many parallel corpora are not publicly accessible for data copyright, data privacy and competitive differentiation reasons, trained translation models are increasingly available on open platforms. In this work, we propose a method called continual knowledge distillation to take advantage of existing translation models to improve one model of interest. The basic idea is to sequentially transfer knowledge from each trained model to the distilled model. Extensive experiments on Chinese-English and German-English datasets show that our method achieves significant and consistent improvements over strong baselines under both homogeneous and heterogeneous trained model settings and is robust to malicious models.\footnote{The source code is available at \url{https://github.com/THUNLP-MT/CKD}.}
\end{abstract}

\section{Introduction}
Current neural machine translation (NMT) systems often face such a situation: parallel corpora are not publicly accessible but trained models are more readily available. On the one hand, many data owners are usually unwilling to share their parallel corpora with the public for data copyright, data privacy and competitive differentiation reasons, leading to recent interests in federated learning for NMT~\cite{privacy, Roosta:21}. On the other hand, trained NMT models are increasingly available on platforms such as Hugginface (\url{https://huggingface.co}) and Opus-MT (\url{https://opus.nlpl.eu/Opus-MT}) since these models can be directly used without public access to the original training data. 

\begin{figure}[!t]
  \centering
  \includegraphics[width=0.47\textwidth]{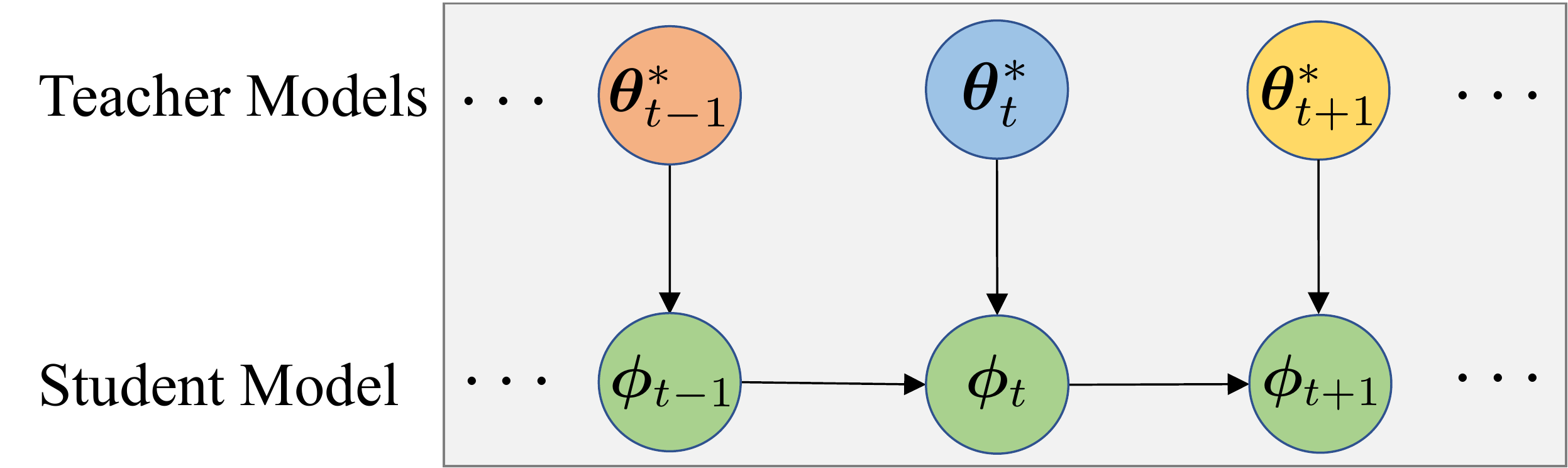}
  \caption{Continual knowledge distillation for neural machine translation. Knowledge is continually distilled from a sequence of teacher models to one student model. At each time step, the current student model (i.e., $\bm{\phi}_{t}$) fuses the knowledge transferred from both the current teacher model (i.e., $\bm{\theta}_t^{*}$) and the previous student model (i.e., $\bm{\phi}_{t-1}$). All teacher models are frozen and the student model is trainable. Different models are highlighted in different colors.}
  \label{fig:pipeline}
\end{figure}

As a result, a question naturally arises: {\em can we take advantage of increasingly available trained NMT models to enhance one NMT model of interest?} In this work, we propose a method called {\bf Continual Knowledge Distillation} (CKD) to address this problem for NMT. As shown in Figure~\ref{fig:pipeline}, we assume that multiple trained NMT models (i.e.,  {\em teachers}) are available to ``educate'' one NMT model of interest (i.e., {\em student}) in a sequential manner, which means that teacher models to arrive in the future are not accessible at the current time step. We also assume that the training set of the student model, a transfer set, and a test set are available, but the training set of the teachers are unavailable. CKD aims to continually improve the translation performance of the student model on the test set by sequentially distilling knowledge from each incoming teacher model to the student model.

As its name suggests, CKD is an intersection of knowledge distillation~\cite{hinton2015distilling} and continual learning~\cite{kirkpatrick2017overcoming}. On the one hand, CKD differs from standard knowledge distillation in that the knowledge is transferred from teacher models to the student model asynchronously instead of synchronously. As a result, the knowledge transferred to the student model from previous teacher models can be overridden by an incoming teacher model, which is often referred to as the {\em catastrophic forgetting} problem~\cite{kirkpatrick2017overcoming}. The situation aggravates when not all teacher models convey knowledge beneficial to the student model. On the other hand, CKD is different from conventional continual learning methods by focusing on learning one task (i.e., enhancing the student model) rather than learning many different tasks. The learning process is still very challenging as compared with standard continual learning because the original training data of teacher models is inaccessible to the student model. Consequently, we have to resort to knowledge distillation at each time step to make the most of teacher models.
 
To address these aforementioned challenges, we propose to fuse two knowledge sources for the student model at each time step: filtering the new knowledge from the current teacher model (i.e., {\em knowledge filtration}) and inheriting the old knowledge from the previous student model (i.e., {\em knowledge inheritance}) simultaneously. Experimental results show that our method significantly and consistently outperforms strong baselines under both homogeneous and heterogeneous teacher settings for Chinese-to-English and German-to-English translation. And it is also robust to malicious teachers.

\section{Approach}

\subsection{Problem Statement}
Let $\bm{\Theta}=\{\bm{\theta}_1^{*}, \bm{\theta}_2^{*}, \dots\}$ be a sequence of frozen trained NMT models (i.e., {\em teacher models}), where $\bm{\theta}_t^{*}$ denotes the $t$-th teacher model. Let $\bm{\phi}_0$ be an NMT model of interest (i.e., {\em student model}) and $\bm{\phi}_t$ be the student model at time step $t$. We use $\mathbf{x} = x_1, \dots, x_I$ to denote a {\em source-language sentence} and $\mathbf{y} = y_1, \dots, y_J$ to denote a {\em target-language sentence}. We use $\mathbf{y}_{<j}=y_1,\dots, y_{j-1}$ to denote a {\em partial translation}. $D_{\mathrm{train}} = \{ \langle \mathbf{x}^{(m)}, \mathbf{y}^{(m)} \rangle\}_{m=1}^{M}$ represents the {\em training set} of the student model. $D_{\mathrm{trans}}=\{ \langle \mathbf{x}^{(n)}, \mathbf{y}^{(n)} \rangle \}_{n=1}^{N}$ represents the {\em transfer set} that a teacher model uses to ``educate'' the student model. $D_{\mathrm{test}}$ is a {\em test set} used to evaluate the student model. We use $\mathrm{BLEU}(D_{\mathrm{test}}, \bm{\phi}_t)$ to denote the BLEU score the student model at time step $t$ obtains on the test set. 

\begin{figure*}[!t]
\begin{center}
\includegraphics[width=0.9\textwidth]{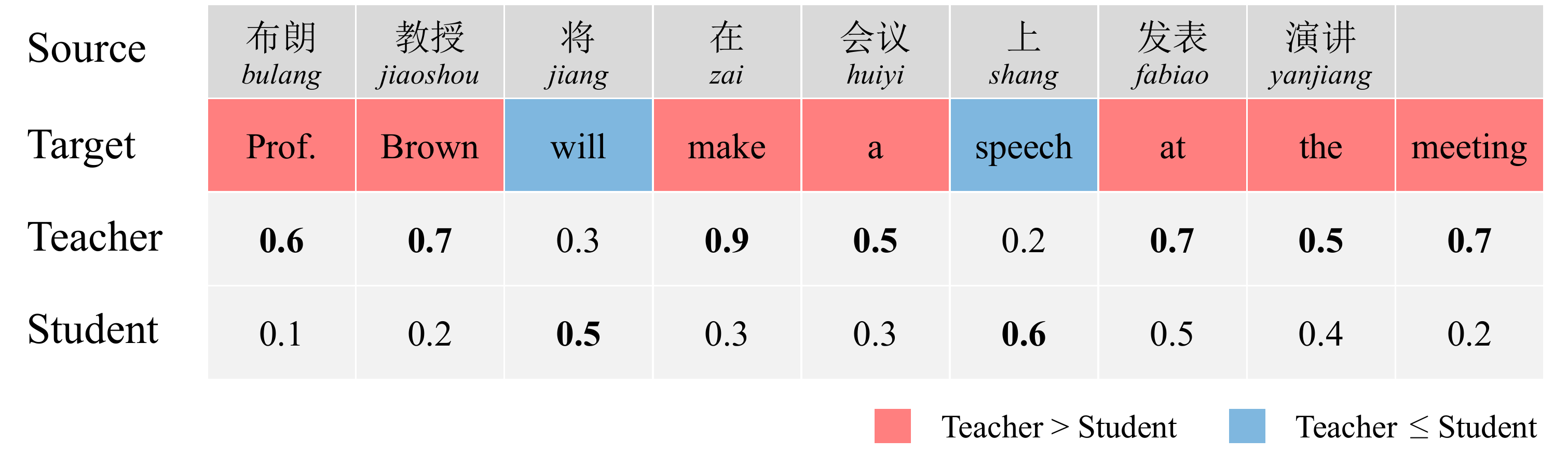}
\caption{An example that illustrates how to find where a teacher model can help a student model. Given a sentence pair of the transfer set, both the teacher and student models try to predict a target word given the source sentence and the partial translation. How well a model predicts can be quantified as a real-valued number. The target words on which the teacher performs better than the student are highlighted in red. Other words are highlighted in blue.} \label{fig:example}
\end{center}
\end{figure*}

Given an initial student model $\bm{\phi}_0$, our goal is to maximize $\mathrm{BLEU}(D_{\mathrm{test}}, \bm{\phi}_t)$ by taking advantage of $\bm{\Theta}$, $D_{\mathrm{train}}$, and $D_{\mathrm{trans}}$.

\subsection{Training Objective}

As shown in Figure \ref{fig:pipeline}, the student model $\bm{\phi}_t$ at time step $t$ is determined by the current teacher model $\bm{\theta}^{*}_t$ that encodes new knowledge and the previous learned student model $\hat{\bm{\phi}}_{t-1}$ that encodes previously learned knowledge. Therefore, the overall training objective of CKD is composed of three loss functions:
\begin{multline}
  \ell(\bm{\phi}_t, \bm{\theta}^{*}_t, \hat{\bm{\phi}}_{t-1}, D_{\mathrm{train}}, D_{\mathrm{trans}})
  \\= \ell_{\mathrm{CE}}(\bm{\phi}_t, D_{\mathrm{train}})
  + \lambda \ell_{\mathrm{KF}}(\bm{\phi}_t, \bm{\theta}^{*}_t, D_{\mathrm{trans}})
  \\+ (1-\lambda) \ell_{\mathrm{KI}}(\bm{\phi}_t, \hat{\bm{\phi}}_{t-1}, D_{\mathrm{trans}}),
  \label{eqn:overall_loss}
\end{multline} 
where $\ell_{\mathrm{CE}}(\bm{\phi}_t, D_{\mathrm{train}})$ is the standard cross entropy loss defined as
\begin{multline}
\ell_{\mathrm{CE}}(\bm{\phi}_t, D_{\mathrm{train}})
\\= - \sum_{m=1}^{M}\sum_{j=1}^{J^{(m)}} P(y^{(m)}_j | \mathbf{y}^{(m)}_{<j}, \mathbf{x}^{(m)}; \bm{\phi}_t),
\end{multline}
Note that $J^{(m)}$ is the length of the $m$-th target sentence $\mathbf{y}^{(m)}$. In Eq.~\ref{eqn:overall_loss}, $\ell_{\mathrm{KF}}(\bm{\phi}_t, \bm{\theta}^{*}_t, D_{\mathrm{trans}})$ is a {\em knowledge filtration} loss (see Sec.~\ref{sec:kf}) that filters the knowledge transferred from $\bm{\theta}^{*}_t$, $\ell_{\mathrm{KI}}(\bm{\phi}_t, \hat{\bm{\phi}}_{t-1}, D_{\mathrm{trans}})$ is a {\em knowledge inheritance} loss (see Sec.~\ref{sec:ki}) that inherits the knowledge transferred from $\hat{\bm{\phi}}_{t-1}$, and $\lambda$ is a hyper-parameter that balances the preference between receiving new and inheriting old knowledge.

Therefore, the learned student model at time step $t$ can be obtained by
\begin{equation}
\hat{\bm{\phi}}_t = \argmin_{\bm{\phi}_t} \Big \{ \ell(\bm{\phi}_t, \bm{\theta}^{*}_t, \hat{\bm{\phi}}_{t-1}, D_{\mathrm{train}}, D_{\mathrm{trans}}) \Big\}.
\end{equation}

\subsection{Knowledge Filtration}
\label{sec:kf}

In standard knowledge distillation~\cite{hinton2015distilling}, an important assumption is that the teacher model is ``stronger'' than the student model, which means that the teacher model contains knowledge that can help improve the student model. Unfortunately, this assumption does not necessarily hold in our problem setting because it is uncertain what the next incoming teacher model will be. As a result, there are two interesting questions:
\begin{enumerate}
\item How do we know whether the teacher model contains knowledge useful to the student model?
\item How do we locate and transfer the useful knowledge from the teacher model to the student model? 
\end{enumerate}

Intuitively, the teacher and the student can do the same ``test paper'' in order to find where the teacher can help the student. Figure \ref{fig:example} shows an example. Given a (Romanized) Chinese sentence and its English translation, both the teacher and student models predict every target word $y_j$ given the source sentence $\mathbf{x}$ and the partial translation $\mathbf{y}_{<y}$. The quality of a prediction can be quantified as a real-valued number. If the teacher model performs better than the student model on a target word (e.g., ``Prof.''), it is likely that the teacher model contains knowledge useful to the student model in this case. On the contrary, the teacher model is probably not more knowledgable than the student model regarding this case if its prediction is worse than that of the student (e.g., ``will'').

More formally, we use $Q(y_j, \mathbf{y}_{<j}, \mathbf{x}, \bm{\phi})$ to quantify how well a student model predicts a target token. It can be defined in the following ways: \footnote{Note that it is also possible to define sentence-level quantification functions $Q(\mathbf{y}, \mathbf{x}, \bm{\phi})$. If the teacher performs better than the student at sentence-level predictions, all target words with the sentence are considered positive instances for knowledge transfer. As our preliminary experiments show that the more fine-grained word-level quantification functions are much better than its sentence-level counterparts, we omit the discussion of sentence-level quantification functions due to the space limit.}

\begin{enumerate}
\item {\em Token entropy:} calculating the entropy of target tokens without using the ground truth token.
\begin{multline}
  Q(y_j, \mathbf{y}_{<j}, \mathbf{x}, \bm{\phi})
  \\= -\sum_{y \in \mathcal{Y}} P(y|\mathbf{y}_{<j}, \mathbf{x}; \bm{\phi})
  \\ \times \log P(y|\mathbf{y}_{<j}, \mathbf{x}; \bm{\phi})
  \label{eqn:token-entropy}
\end{multline}
where $\mathcal{Y}$ is the vocabulary of the target language.

\item {\em Hard label matching}: checking whether the predicted token is identical to the ground truth.
\begin{multline}
  Q(y_j, \mathbf{y}_{<j}, \mathbf{x}, \bm{\phi})
  \\= \delta\Big(y_j, \argmax_{y}P(y|\mathbf{y}_{<j}, \mathbf{x}; \bm{\phi}) \Big)
  \label{eqn:hard-label-matching}
\end{multline}
where $\delta(y, y')$ returns 1 if $y$ is identical to $y'$ and 0 otherwise. 

\item {\em Token-level cross entropy: } calculating token-level cross entropy using the given model.
\begin{equation}
  Q(y_j, \mathbf{y}_{<j}, \mathbf{x}, \bm{\phi})  = -\log P(y_j | \mathbf{y}_{<j}, \mathbf{x}; \bm{\phi})
  \label{eqn:token-level-ce}
\end{equation}
\end{enumerate}
The quantification function for a teacher model $Q(y_j, \mathbf{y}_{<j}, \mathbf{x}, \bm{\theta})$ can be defined likewise.

Since the transfer set $D_{\mathrm{trans}}$ can be equivalently seen as a collection of tuples
\begin{equation}
  \Big\{ \big\langle y^{(n)}_j, \mathbf{y}^{(n)}_{<j}, \mathbf{x}^{(n)} \big \rangle \Big | j \in [1, J^{(n)} ] , n \in [1, N]  \Big\},
\end{equation}
it can be divided into two parts depending on the comparison between the predictions of teacher and student models: a {\em positive subset} $D^{+}_{\mathrm{trans}}$ and a {\em negative subset} $D^{-}_{\mathrm{trans}}$. A tuple $\langle y_j, \mathbf{y}_{<j}, \mathbf{x} \rangle$ belongs to $D^{+}_{\mathrm{trans}}$ if $Q(y_j, \mathbf{y}_{<j}, \mathbf{x}, \bm{\theta}^{*}_t)$ is greater than $Q(y_j, \mathbf{y}_{<j}, \mathbf{x}, \bm{\phi}_t)$. Otherwise, it is a negative instance that belongs to $D^{-}_{\mathrm{trans}}$.

After splitting the transfer set into two parts, it is natural to apply standard knowledge distillation using the positive subset $D^{+}_{\mathrm{trans}}$:
\begin{multline}
  \ell_{\mathrm{KD}}(\bm{\phi}_t, \bm{\theta}^{*}_t, D^{+}_{\mathrm{trans}})
  \\ = \sum_{\langle y_j, \mathbf{y}_{<j}, \mathbf{x} \rangle \in D^{+}_{\mathrm{trans}}} \mathrm{KL}\Big(P(y_j|\mathbf{y}_{<j}, \mathbf{x}; \bm{\theta}^{*}_t) \big|\big| 
  \\P(y_j|\mathbf{y}_{<j}, \mathbf{x}; \bm{\phi}_t) \Big) 
\end{multline}
However, one problem is that $D^{+}_{\mathrm{trans}}$ may be very small in most cases in practice, making training efficiency very low.

Therefore, instead of discarding the negative subset $D^{-}_{\mathrm{trans}}$, we introduce a new loss function to make the most of negative instances. In analogy to humans, teachers can educate students by telling them what not to do. We expect that the student model can learn from $D^{-}_{\mathrm{trans}}$ in the same way. Our intuition is that erroneous tokens with a high probability in teacher model’s output distribution are critical because the student is prone to make the same mistakes. Pushing the output distribution of the student model away from the poor target distribution may enable the student model to avoid making the same mistakes. As a result, $D^{-}_{\mathrm{trans}}$ can be leveraged effectively and the overall learning efficiency will be improved significantly. Accordingly, the negative KD loss function on the negative subset is defined as
\begin{multline}
  \ell_{\mathrm{NEG}}(\bm{\phi}_t, \bm{\theta}^{*}_t, D^{-}_{\mathrm{trans}})
  \\= \min \Big(0, \alpha - \ell_{\mathrm{KD}}(\bm{\phi}_t, \bm{\theta}^{*}_{t}, D^{-}_{\mathrm{trans}}) \Big)
  \label{eqn:neg-kd}
\end{multline}
where $\alpha$ is a hyper-parameter that controls the activation of the loss.

Finally, the knowledge filtration loss is the combination of the two functions:
\begin{multline}
\ell_{\mathrm{KF}}(\bm{\phi}_t, \bm{\theta}^{*}_t, D_{\mathrm{trans}})
\\ = \ell_{\mathrm{KD}}(\bm{\phi}_t, \bm{\theta}^{*}_t, D^{+}_{\mathrm{trans}}) +  \ell_{\mathrm{NEG}}(\bm{\phi}_t, \bm{\theta}^{*}_t, D^{-}_{\mathrm{trans}}).
\label{eqn:sum_all}
\end{multline}

\subsection{Knowledge Inheritance} \label{sec:ki}
To circumvent the catastrophic forgetting problem, we introduce a loss function to inherit knowledge learned from previous time step for the current student model: 
\begin{multline}
  \ell_{\mathrm{KI}}(\bm{\phi}_t, \hat{\bm{\phi}}_{t-1}, D_{\mathrm{trans}})
  \\ = \sum_{\langle y_j, \mathbf{y}_{<j}, \mathbf{x} \rangle \in D_{\mathrm{trans}}} \mathrm{KL} \Big( P(y_j | \mathbf{y}_{<j}, \mathbf{x}; \hat{\bm{\phi}}_{t-1} \big|\big|
  \\P(y_j | \mathbf{y}_{<j}, \mathbf{x}; \bm{\phi}_{t} ) \Big).
\end{multline}

\section{Experiments}

\begin{table}
\centering
\small
    \begin{threeparttable}
    \begin{tabular}{clrrr}
    \toprule
    \textbf{Model} & \textbf{Domain} & \textbf{Training} & \textbf{Dev.} &
    \textbf{Test}  \\
    \midrule
    $A$ &  News & 1,250,000  & 4,000 & 13,000
     \\
    $B$ &  Oral & 2,500,000 & 4,000 & 12,000
    \\
    $C$ &  Internet & 750,000 & 4,000 & 13,000
    \\ 
    $D$ &  Speech & 220,000 & 4,000& 5,000
    \\
    $E$ &  Subtitle & 300,000 & 4,000 & 4,000  \\
    \bottomrule
    \end{tabular}
    \end{threeparttable}
    \captionof{table}{The domain, training and evaluation corpora of the five Transformer-base models used in the Chinese-to-English experiments. More details of the datasets are provided in Appendix~\ref{sec:appendix:datasets}.
    }
    \label{tab:corpora}
\end{table}

To evaluate the effectiveness of our method, we conduct experiments on Chinese-to-English and  German-to-English translation under three representative settings including homogeneous, heterogeneous and malicious teacher settings.

\begin{table*}[!t]
  \centering
  \begin{threeparttable}
    \small
    \begin{tabular}{cl|lrlrlr|lr} 
    \toprule
    \multirow{2}[3]{*}{\bf Step}&\multirow{2}[3]{*}{\bf Method}&
    \multicolumn{2}{c}{\bf BCDE$\rightarrow$A}&\multicolumn{2}{c}{\bf ACDE$\rightarrow$B}&\multicolumn{2}{c|}{\bf ABDE$\rightarrow$C}&\multicolumn{2}{c}{\bf Average}\\
    \cmidrule(lr){3-4} \cmidrule(lr){5-6} \cmidrule(lr){7-8}
    \cmidrule(lr){9-10}
    && \multicolumn{1}{c}{{\bf BLEU$\mathbf{\uparrow}$}} & \multicolumn{1}{c}{{\bf AD$\mathbf{\downarrow}$}} & \multicolumn{1}{c}{{\bf BLEU$\mathbf{\uparrow}$}} & \multicolumn{1}{c}{{{\bf AD$\mathbf{\downarrow}$}}} & \multicolumn{1}{c}{{\bf BLEU$\mathbf{\uparrow}$}} & \multicolumn{1}{c|}{{\bf AD$\mathbf{\downarrow}$}} & \multicolumn{1}{c}{{\bf BLEU$\mathbf{\uparrow}$}} & \multicolumn{1}{c}{{\bf AD$\mathbf{\downarrow}$}} \\
    \midrule\midrule
    0 & & 42.84  & & 27.53  & & 18.06 & & 29.48 & /
    \cr
    \midrule

    \multirow{4}{*}{1} & KD & 46.19$_{\text{3.35}}$ & 0.00 & 24.32$_{\text{-3.21}}$ & 3.21 & 17.06$_{\text{-1.00}}$ & 1.00 & 29.19$_{\text{-0.29}}$ & 1.40

    \cr
    & EWC & 46.09$_{\text{3.25}}$ & 0.00 & 24.32$_{\text{-3.21}}$ & 3.21 & 17.12$_{\text{-0.94}}$ & 0.94 & 29.18$_{\text{-0.30}}$ & 1.38
    \cr
    & CL-NMT & 46.14$_{\text{3.30}}$ & 0.00 & 24.28$_{\text{-3.25}}$ & 3.25 & 17.09$_{\text{-0.97}}$ & 0.97 & 29.17$_{\text{-0.31}}$ & 1.41
    
    \cr
    & Ours & 46.00$_{\text{3.16}}$ & 0.00 & 28.01$_{\text{0.48}}$ & 0.00 & 18.98$_{\text{0.92}}$ & 0.00 & 31.00$_{\text{1.52}}$ & 0.00
    \cr
    \midrule
    \multirow{4}{*}{2} & KD & 44.62$_{\text{1.78}}$ & 1.57 & 26.11$_{\text{-1.42}}$ & 3.21 & 19.33$_{\text{1.27}}$ & 1.00 & 30.02$_{\text{0.54}}$ & 1.93
    \cr
    & EWC & 45.80$_{\text{2.96}}$ & 0.29 & 25.28$_{\text{-2.25}}$ & 3.21 & 18.13$_{\text{0.07}}$ & 0.94 & 29.74$_{\text{0.26}}$ & 1.48
    \cr
    & CL-NMT & 45.24$_{\text{2.40}}$ & 0.90 & 27.67$_{\text{0.14}}$ & 3.25 & 19.09$_{\text{1.03}}$ & 0.97 &  30.67$_{\text{1.19}}$ & 1.71
    
    \cr
    & Ours & 45.89$_{\text{3.05}}$ & 0.11 & 28.28$_{\text{0.75}}$ & 0.00 & 19.18$_{\text{1.12}}$ & 0.00 & 31.12$_{\text{1.64}}$ & 0.04
    \cr
    \midrule
    \multirow{4}{*}{3} & KD & 39.16$_{\text{-3.68}}$ & 7.03 & 21.76$_{\text{-5.77}}$ & 7.56 & 16.14$_{\text{-1.92}}$ & 4.19 & 25.69$_{\text{-3.79}}$ & 6.26
    \cr
    & EWC & 43.88$_{\text{1.04}}$ & 2.21 & 24.48$_{\text{-3.05}}$ & 4.01 & 17.76$_{\text{-0.30}}$ & 1.31 & 28.71$_{\text{-0.77}}$ & 2.51
    \cr
    
    & CL-NMT & 43.91$_{\text{1.07}}$ & 2.23 & 27.23$_{\text{-0.3}}$ & 3.69 & 18.45$_{\text{0.39}}$ & 1.61 & 29.86$_{\text{0.39}}$ & 2.51
    
    \cr
    & Ours & 45.89$_{\text{3.05}}$ & 0.11 & 28.41$_{\text{0.88}}$ & 0.00 & 19.15$_{\text{1.09}}$ & 0.03 & 31.15$_{\text{1.67}}$ & 0.05
    \cr
    \midrule
    \multirow{4}{*}{4} & KD & 30.57$_{\text{-12.27}}$ & 15.62 & 22.71$_{\text{-4.82}}$ & 7.56 & 13.88$_{\text{-4.18}}$ & 6.45 & 22.39$_{\text{-7.09}}$ & 9.88
    \cr
    & EWC & 41.13$_{\text{-1.71}}$ & 4.96 & 24.89$_{\text{-2.64}}$ & 4.01 & 17.24$_{\text{-0.82}}$ & 1.83 & 27.75$_{\text{-1.72}}$ & 3.60
    \cr
    & CL-NMT & 43.13$_{\text{0.29}}$ & 3.01 & 27.90$_{\text{0.37}}$ & 3.69 & 18.59$_{\text{0.53}}$ & 1.61 & 29.87$_{\text{0.40}}$ & 2.77
    
    \cr
    & Ours & \textbf{45.89}$_{\text{3.05}}$ & \textbf{0.11} & \textbf{28.49}$_{\text{0.96}}$ & \textbf{0.00} & \textbf{19.15}$_{\text{1.09}}$ & \textbf{0.03} & \textbf{31.18}$_{\text{1.70}}$ & \textbf{0.05}
    \cr
    \bottomrule
    \end{tabular}
    \end{threeparttable}
    \caption{Results of \textit{Chinese-to-English} translation under \textit{homogeneous} teacher setting. ``BCDE$\rightarrow$A'' denotes $A$ is the student model and $B$, $C$, $D$, and $E$ are teacher models in step $1$ to $4$,  respectively. ``$\Delta$'' denotes $\Delta$BLEU compared with step $0$ (i.e., initial student model), and $\Delta$BLEU scores are also reported as subscript numbers. ``AD'' is the accumulative degradation defined in Eq.~\ref{eqn:AQD}, which is the lower the better. The last two columns are numbers averaged row-wise. Best results in step $4$ are in {\bf bold}.}
    \label{tab:main2}
\end{table*}

\subsection{Setup}
\paragraph{Configurations.}
For the Chinese-to-English translation experiments under the homogeneous teacher setting, both the teachers and the student are Transformer-base models~\cite{vaswani2017attention}.
Besides model architecture, there are a few other factors that may affect performance, e.g.,  teacher performance, student performance, model domain, and the order that the teachers arrive.
To investigate the impact of model performance and model domain, we leverage five parallel corpora of representative domains as shown in Table~\ref{tab:corpora}, among which two are in million scale, one is in middle scale, and the other two are in small scale. Correspondingly, five Transformer-base models are trained on these corpora, denoted as $A$, $B$, $C$, $D$, and $E$, respectively. Intuitively, $A$ and $B$ are well-trained while $D$ and $E$ are under-trained due to the training data sizes.
To investigate the impact of the order of teachers, we enumerate all the six permutations of $A$, $B$ and $C$. In addition, we append $D$ and $E$ to the end of each permutation to simulate the ``weak'' teacher scenario. Therefore, we have six configurations in total.

Specially, we use a string like ``ABDE $\rightarrow$ C'' to denote a configuration, which means $C$ is the student, $A$, $B$, $D$ and $E$ are the teachers and $A$ arrives first, then $B$ and so on. For simplicity, we use the training set of $C$ as both the training set $D_{\mathrm{train}}$ and the transfer set $D_{\mathrm{trans}}$ in CKD, and the test set of $C$ is leveraged as $D_{\mathrm{test}}$. The goal in this configuration is to improve the performance of $C$ on $D_{\mathrm{test}}$. In summary, the six configurations are ``BCDE$\rightarrow$A'', ``CBDE$\rightarrow$A'', ``ACDE$\rightarrow$B'', ``CADE$\rightarrow$B'', ``ABDE$\rightarrow$C'', and ``BADE$\rightarrow$C''.

For clarity, the differences of other aforementioned settings with this one  will be given in the corresponding sections later.

\paragraph{Evaluation.} We leverage the following two metrics to evaluate our method:
\begin{itemize}
    \setlength{\itemsep}{0em}
    \item \textit{BLEU}~\cite{bleu2002papineni} \footnote{BLEU score is computed using \texttt{multi-bleu.perl}
    on the corresponding test set for each student model.}: the most widely used evaluation metric for machine translation.

    \item \textit{Accumulative Degradation} (AD): measuring the accumulative occasional quality degradation in all steps, which should be avoided as much as possible. AD from step $1$ to $t$ is defined as follows:
    \vspace{-0.5em}
    \begin{equation}
        \mathrm{AD}=\sum_{k=1}^t \max(0, \mathrm{B}(\bm{\phi}_{k-1})-\mathrm{B}(\bm{\phi}_k)),
        \label{eqn:AQD}
    \end{equation}
    where $\mathrm{B}(\cdot)$ denotes $\mathrm{BLEU}(D_{\mathrm{test}}, \cdot)$.
\end{itemize}

\paragraph{Baselines.} Our method is compared with the following baseline methods:
\begin{itemize}
    \setlength{\itemsep}{0em}
    \item {\it Knowledge Distillation} (KD)~\cite{khayrallah-etal-2018-regularized} for NMT which applies vanilla knowledge distillation on each token trivially.
    \item {\it Elastic Weight Consolidation} (EWC)~\citep{saunders2019domain,thompson-etal-2019-overcoming} which is a representative continual learning method that adds an EWC term as a penalty to alleviate catastrophic forgetting.
    \item {\it Continual Learning for NMT} (CL-NMT) \cite{cao2021continual} which is a representative work on \emph{multi-step} continual learning in NMT.
\end{itemize}

\subsection{Implementation Details} 
We use byte pair encoding (BPE)~\cite{sennrich2015neural} with the vocabulary size of $32$k. 
The hyper-parameters of the Transformer-base models are set mostly following \citet{vaswani2017attention}. We use Adam~\cite{kingma2014adam} optimizer, in which $\beta_1=0.9$, $\beta_2=0.98$. During training, learning rate is $7\times 10^{-4}$ and dropout rate is $0.1$. 
Batch size is $6,000$. $\lambda$ in Eq.~\ref{eqn:overall_loss} in step $t$ is defined as $\lambda=0.999 \frac{1-0.999^{t-1}}{1-0.999^{t}}$ following~\cite{cao2021continual}. More details of hyper-parameters are provided in Appendix~\ref{sec:appendix:hyper-parameter}.

\subsection{Quantification Function Selection}

We first evaluate the three candidates for the quantification function $Q$ defined in Sec.~\ref{sec:kf}. 
A proper $Q$ should correlate well with model performance and generalize well to a wide range of domains.
To this end, we collect six widely used datasets of different domains and varying sizes and evaluate the correlations between the candidates and corpus-level BLEU scores on them. The Pearson correlation coefficients between token entropy (Eq.~\ref{eqn:token-entropy}), hard label matching (Eq.~\ref{eqn:hard-label-matching}) and token-level cross entropy (Eq.~\ref{eqn:token-level-ce}) are $-0.5622$, $0.8091$ and $0.7792$, respectively.
Both hard label matching and token-level cross entropy are strongly correlated with corpus-level BLEU. However, hard label matching can not break a tie when both the teacher and student models' predictions are correct or incorrect. 
Therefore, we adopt token-level cross entropy as $Q$ in the rest of this work.
Examples and more discussions can be found in Appendix~\ref{appendixmetric}.

\subsection{Chinese-to-English Translation}
\label{sec:zh-en-translation}
\paragraph{Homogeneous Teacher Setting.} In this setting, all the student and teacher models are of the same model architecture, which is Transformer-base. 
For space limitation, we only show results of three configurations in Table~\ref{tab:main2}. The full results for all configurations can be found in Appendix~\ref{full}. From Table~\ref{tab:main2} we can observe that:

(1) Our method achieves improvements over the initial student model in all steps and configurations, and outperforms all baselines significantly. 
It indicates that our method is effective for leveraging diverse teacher models to continually improve the performance of the student model on its test dataset.

(2) Our method achieves zero or near-zero accumulative performance degradation (AD) scores in all configurationss, indicating our method is also effective to retain acquired knowledge.
Especially, when encountering model $D$ (step $3$), nearly all baselines face severe quality degradation compared with step $2$, while our method even achieves gain in $ACDE\rightarrow B$, which further justifies the effectiveness of our method.

(3) All baselines perform poorly after four steps of distillation, indicating that the problem we aim to resolve is challenging. Specifically, KD, the worst one, suffers from severe performance degradation as averaged $\Delta$BLEU and AD scores are $-7.09$ and $9.88$, respectively. We argue this is due to KD implicitly assumes that the teacher models are helpful such that it is prone to less beneficial knowledge provided by them. EWC is designed to alleviate catastrophic forgetting and achieves better $\Delta$BLEU and AD scores than KD. However, EWC still fails to achieve improvement over the initial student model, i.e., all $\Delta$BLEU scores are negative. CL-NMT is specially designed for multi-step continual learning in NMT and achieves the best $\Delta$BLEU and AD scores among baselines. Nevertheless, its average $\Delta$BLEU score is significantly smaller than ours ($0.40$ v.s. $1.70$) and its average AD score is significantly worse than ours ($2.77$ v.s. $0.05$). Overall, the problem to be resolved is challenging and our method is remarkably effective than baselines.

(4) Despite the promising results,  slight performance degradation can still be observed occasionally for our method. Therefore, there is still room for further improvement on retaining acquired knowledge.

\begin{table}[!t]
\centering
\resizebox{0.48\textwidth}{!}{
\begin{threeparttable}
        \small\setlength\tabcolsep{3pt}
        \begin{tabular}{l|lrlrlr}
        \toprule
        \multirow{2}[3]{*}{\bf
        Method}&
        \multicolumn{2}{c}{\bf Base$\rightarrow$Base}&\multicolumn{2}{c}{\bf RNN$\rightarrow$Base}&\multicolumn{2}{c}{\bf Big$\rightarrow$Base} \\
        \cmidrule(lr){2-3} \cmidrule(lr){4-5} \cmidrule(lr){6-7}
        &\multicolumn{1}{c}{\bf {BLEU}$\mathbf{\uparrow}$ }&\multicolumn{1}{c}{\bf{AD}$\mathbf{\downarrow}$} &\multicolumn{1}{c}{\bf {BLEU}$\mathbf{\uparrow}$ }&\multicolumn{1}{c}{\bf{AD}$\mathbf{\downarrow}$} &\multicolumn{1}{c}{\bf {BLEU}$\mathbf{\uparrow}$ }&\multicolumn{1}{c}{\bf{AD}$\mathbf{\downarrow}$}    \\
        \midrule\midrule
        Original & 29.48 & / & 29.48 & / & 29.48 & /
        \\
        \midrule
         KD  & 29.58$_{\text{0.09}}$ & 0.93 & 28.29$_{\text{-1.19}}$ & 1.21 & 29.21$_{\text{-0.27}}$ & 1.17
         \cr
         EWC & 29.63$_{\text{0.15}}$ & 0.90 & 27.83$_{\text{-1.65}}$ & 1.65 & 29.39$_{\text{-0.09}}$ & 0.32
        \cr
        CL-NMT & 29.57$_{\text{0.09}}$ & 0.92 & 25.89$_{\text{-3.59}}$ & 3.59  & 28.00$_{\text{-1.48}}$ & 1.91
        
        \cr
        
         Ours & \textbf{31.07}$_{\text{1.59}}$ &  \textbf{0.00} &  \textbf{29.44}$_{\text{-0.04}}$ & \textbf{0.12} &  \textbf{30.82}$_{\text{1.34}}$ & \textbf{0.00}
        \cr
        \bottomrule
        \end{tabular}
        \end{threeparttable}}
        \captionof{table}{Results of \textit{Chinese-to-English} translation under the \textit{heterogeneous} teacher setting in step $1$, averaged over six configurations. ``Base'' and ``Big'' denote Transformer-based and Transformer-big models, respectively. And ``X$\rightarrow$Y'' denotes that X is the teacher and Y is the student.}
        \label{tab:arc}
\end{table}

\begin{table}[!t]
\centering
\resizebox{0.48\textwidth}{!}{
\begin{threeparttable}
        \small\setlength\tabcolsep{3pt}
        \begin{tabular}{l|lrlrlr}
        \toprule
        \multirow{2}[3]{*}{\bf
        Method}&
        \multicolumn{2}{c}{\bf Base (M)$\rightarrow$Base}&\multicolumn{2}{c}{\bf RNN (M)$\rightarrow$Base}&\multicolumn{2}{c}{\bf Big (M)$\rightarrow$Base} \\
        \cmidrule(lr){2-3} \cmidrule(lr){4-5} \cmidrule(lr){6-7}
        &\multicolumn{1}{c}{\bf {BLEU$\mathbf{\uparrow}$} }&\multicolumn{1}{c}{\bf{AD}$\mathbf{\downarrow}$} &\multicolumn{1}{c}{\bf {BLEU}$\mathbf{\uparrow}$ }&\multicolumn{1}{c}{\bf{AD}$\mathbf{\downarrow}$} &\multicolumn{1}{c}{\bf {BLEU}$\mathbf{\uparrow}$}&\multicolumn{1}{c}{\bf{AD}$\mathbf{\downarrow}$}    \\
        \midrule\midrule
        Original & 29.48 & / & 29.48 & / & 29.48 & /
        \\
        \midrule
         KD  & 18.36$_{\text{-11.12}}$ & 11.12 & 13.88$_{\text{-15.60}}$ & 15.60 & 18.53$_{\text{-10.95}}$ & 10.95
         \cr
        
         EWC & 24.13$_{\text{-5.35}}$ & 5.35 & 22.84$_{\text{-6.64}}$ & 6.64 & 23.04$_{\text{-6.44}}$ & 6.44
         \cr
        CL-NMT & 11.14$_{\text{-18.34}}$ & 18.34 & {\color{white}0}3.16$_{\text{-26.32}}$ & 26.32 & 10.90$_{\text{-18.58}}$ & 18.58

        \cr
         Ours &  \textbf{29.48}$_{\text{0.00}}$ & \textbf{0.00} &\textbf{29.48}$_{\text{0.00}}$ & \textbf{0.00} &\textbf{29.48}$_{\text{0.00}}$ &\textbf{0.00}
        \cr
        \bottomrule
        \end{tabular}
        \end{threeparttable}}
        \captionof{table}{Results of \textit{Chinese-to-English} translation under the \emph{malicious} teacher setting in step $1$, averaged over six configurations. ``(M)'' is short for ``malicious''.}
        \label{tab:malicious}
\end{table}

\paragraph{Heterogeneous Teacher Setting.} 
Using logits as the medium to transfer and retain knowledge, our approach is model-agnostic and scalable. To justify that, we replace the Transformer-base teacher models with RNN~\cite{bahdanau2014neural} and Transformer-big~\cite{vaswani2017attention} models, and repeat the experiments in Table~\ref{tab:main2} with other settings remaining identical. 
Table~\ref{tab:arc} shows similar results as Table~\ref{tab:main2} that our method outperforms all baselines significantly and also achieves zero or near-zero AD scores, indicating that our method is extensible to different model architectures.
Interestingly, all the baselines encounter serious performance degradation while the $\Delta$BLEU of our method is nearly zero, indicating that distilling knowledge from a teacher of a completely different architecture may be extremely difficult. It deserves more thoughtful investigation and we leave it as future work.

\paragraph{Malicious Teacher Setting.} 
Robustness to malicious models is critical in our scenario as only the parameters rather than training data of teachers are available. We simulate malicious teacher models by shuffling the outputs of a well-trained model within a batch so that the model answers almost completely wrong with high confidence. We repeat the experiments in Table~\ref{tab:main2} with other settings remaining identical. As shown in Table~\ref{tab:malicious}, our approach is far less affected by the malicious model with three different teacher model architectures. Moreover, it could be further explored to detect and skip malicious models to save computational resources directly. 

\subsection{Larger Scale Chinese-to-English Translation}
We scale up the dataset size of the Chinese-to-English translation experiment under the homogeneous teacher setting from one million to ten million. Other settings are similar to the original experiments and are detailed in Appendix~\ref{app:larger}. As shown in Table~\ref{tab:bigger_main}, our method remains effective while all baseline methods fail to achieve positive quality gain ($\Delta$BLEU). This demonstrates that the performance of the baseline methods does not improve as the size of the data and performance of the models increase, while our method remains valid. Thus, it shows that our method is scalable for corpus of different sizes.

\begin{table}[!t]
  \centering
  \small
  \begin{threeparttable}
    \begin{tabular}{l|lr}
    \toprule
    {\bf Method} & \multicolumn{1}{c}{\bf BLEU$\mathbf{\uparrow}$ }  & \multicolumn{1}{c}{\bf AD$\mathbf{\downarrow}$}
    \cr
    \midrule
    \midrule
    Original & 32.77 & /
    \cr
    \midrule
     KD & 28.79$_{\text{-3.98}}$   & 7.95
     \cr
     EWC & 31.54$_{\text{-1.23}}$    & 2.56
     \cr
     CL-NMT & 31.01$_{\text{-1.76}}$  & 3.42
    
    \cr
    Ours & \textbf{33.43}$_{\text{ 0.66}}$  & \textbf{0.03}
    \cr
    \bottomrule
    \end{tabular}
    \end{threeparttable}
      \captionof{table}{Results of extending the training data of the \emph{Chinese-to-English} teacher models to ten million scale under the \emph{homogeneous} teacher setting, averaged over six configurations.}
        \label{tab:bigger_main}
\end{table}

\subsection{German-to-English Translation}

We also conduct experiments on German-to-English datasets. Models are trained on four different datasets from different domains. Other settings are similar to the Chinese-to-English experiments and are detailed in Appendix~\ref{app:ende}. The average values among each of the homogeneous, heterogeneous, and malicious teacher settings are reported in Table~\ref{tab:ende_main}. Due to the large domain differences of the datasets, only our method consistently obtains BLEU gains and zero or near zero AD scores, exceeding the baselines, demonstrating that our approach is effective for different language pairs.

\subsection{Ablation Study}
Table~\ref{tab:sub2} shows the effect of the negative KD loss $\ell_\mathrm{NEG}$ (Eq.~\ref{eqn:neg-kd}) in knowledge filtration and the knowledge inheritance loss $\ell_\mathrm{KI}$. Results at the beginning ($t=1$) and later step ($t=4$) for Chinese-to-English translation under the homogeneous teacher setting are reported. We can observe that: 
\begin{enumerate}
    \item Removing either $\ell_\mathrm{NEG}$  (row 2) or $\ell_\mathrm{KI}$ (row 4)  hurts the performance, indicating both of them are effective.
    \item Comparing row 1 with row 2, we can conclude that the negative subset of the transfer set where the teacher performs worse than the student ($D_{\mathrm{trans}}^-$) also contains valuable nontrivial knowledge. Furthermore, trivially applying vanilla KD loss $\ell_\mathrm{KD}$ on $D_{\mathrm{trans}}^-$ (row 2 v.s. 3) brings no gain. Therefore, our proposed negative KD loss is effective for making less beneficial knowledge play a good role.
    \item Without $\ell_\mathrm{KI}$, the performance drops severely,  especially at a later step, verifying that knowledge inheritance is essential for retaining acquired knowledge. 
\end{enumerate}

\begin{table}[!t]
\centering
\resizebox{0.48\textwidth}{!}{
\begin{threeparttable}
        \small\setlength\tabcolsep{3pt}
        \begin{tabular}{l|lrlrlr}
        \toprule
        \multirow{2}[3]{*}{\bf
        Method}&
        \multicolumn{2}{c}{\bf Homogeneous}&\multicolumn{2}{c}{\bf Heterogeneous}&\multicolumn{2}{c}{\bf Malicious } \\
        \cmidrule(lr){2-3} \cmidrule(lr){4-5} \cmidrule(lr){6-7}
        &\multicolumn{1}{c}{\bf {BLEU}$\mathbf{\uparrow}$ }&\multicolumn{1}{c}{\bf{AD}$\mathbf{\downarrow}$} &\multicolumn{1}{c}{\bf {BLEU}$\mathbf{\uparrow}$ }&\multicolumn{1}{c}{\bf{AD}$\mathbf{\downarrow}$} &\multicolumn{1}{c}{\bf {BLEU}$\mathbf{\uparrow}$ }&\multicolumn{1}{c}{\bf{AD}$\mathbf{\downarrow}$}    \\
        \midrule\midrule
        Original & 30.62 & / & 30.62 & / & 30.62 & /
        \\
        \midrule
         KD	 & 30.68$_{\text{0.06}}$	 & 0.13 & 	30.19$_{\text{-0.43}}$	& 0.44	 & 17.87$_{\text{-12.75}}$ & 	12.75
          \cr
         EWC & 	30.66$_{\text{0.04}}$	 & 0.18 & 30.39$_{\text{-0.23}}$	 & 0.25	 & 24.26$_{\text{-6.36}}$ & 	6.36

\cr
        CL-NMT & 	30.85$_{\text{0.23}}$ & 	0.10	 & 26.20$_{\text{-4.42}}$ & 	4.42 & 	{\color{white}0}8.49$_{\text{-22.1}}$	 & 22.13

        \cr
         Ours & 	\textbf{31.19}$_{\text{0.57}}$ & 		 \textbf{0.00}	 & 	 \textbf{30.85}$_{\text{0.23}}$	 & 	 \textbf{0.05} & 		\textbf{30.62}$_{\text{0.00}}$	 & 	\textbf{0.00}

        \cr
        \bottomrule
        \end{tabular}
        \end{threeparttable}}
        \captionof{table}{Results of \textit{German-to-English} translation in step $1$, averaged over all six setting groups.}
        \label{tab:ende_main}
\end{table}

\begin{table}[!t]
  \centering
  \small
  \begin{threeparttable}
    \renewcommand\tabcolsep{3.0pt}
    \begin{tabular}{llrr}
    \toprule
    &{\bf Method} & \multicolumn{1}{c}{\bf Step 1} & \multicolumn{1}{c}{\bf Step 4}
    \cr
    \midrule
    1 & Full Model & {\bf 31.07} & {\bf 31.18}
    \cr
    2 & \quad Removing $\ell_\mathrm{NEG}$ & 30.69 & 30.54
    \cr
    3 & \quad Replacing $\ell_\mathrm{NEG}$ with $\ell_\mathrm{KD}$& 30.60 & 30.31 
    \cr
    4 & \quad Removing $\ell_\mathrm{KI}$ & 30.74 & 29.94
    \cr
    \bottomrule
    \end{tabular}
    \end{threeparttable}
      \caption{Ablation study on \emph{Chinese-to-English} translation under \emph{homogeneous} teacher setting. BLEU scores averaged over six configurations are reported.}

      \label{tab:sub2}
\end{table}

\subsection{Comparison with Multi-teacher Knowledge Distillation}
\label{sec:ensem}
Multi-teacher KD~\cite{Freitag-2017-ensemble-kd}, aka ensemble KD, generally requires all teachers available at the same time, which violates the definition of our problem and may result in  enormous computational and memory cost as teacher number grows. Moreover, it is also non-trivial to adapt it to our scenarios due to potential unbeneficial knowledge provided by teachers. Therefore, we do not include it as a major baseline in the experiments above. Nevertheless, in this section, we still provide a comparison of our method with vanilla multi-teacher KD which averages the outputs of all teachers as the target distribution for analysis. 
The BLEU score of vanilla multi-teacher KD averaged over six configurations is $30.49$, lower than our $31.18$, indicating that our method is superior to vanilla multi-teacher KD although  the comparison is more favorable to it. 
More details on comparison in terms of task definition, robustness and storage requirement are analyzed in Appendix~\ref{app:ensemble}.

\section{Related Work}
\paragraph{Knowledge Distillation.} Knowledge distillation (KD) is the most widely used technique for transferring knowledge between models~\citep{hinton2015distilling}. Despite of their effectiveness, conventional KD methods usually implicitly assume that the teacher model is superior or complementary to the student model~\cite{gou2021knowledge}. Although recently \citet{qin2021knowledge} allow a big model to learn from small models, they still require that the small models are better than the big model for the given tasks and datasets. However, the assumption does not necessarily hold in our scenario due to the diversity of teacher models. Multi-teacher KD~\citep{Freitag-2017-ensemble-kd,you2017learning,fukuda2017efficient,mirzadeh2020improved,liu2020adaptive}, which distills knowledge from multiple teachers simultaneously, is highly related to this work. Generally, multi-teacher KD requires all teachers to be available at the same time, which will result in enormous extra memory consummation as the number of teachers grows. More importantly, new teachers may be released constantly~\cite{wolf-etal-2020-transformers}, which can not be seen in advance. Therefore, multi-teacher KD methods are not feasible to our scenario. L2KD~\cite{chuang-etal-2020-lifelong-L2KD} leverages sequential KD to continually learn new tasks, having different goal and challenges compared with our scenario. Another line of related work is selective distillation~\citep{gu_train_2020,wang_selective_2021,shi_data_2022}, which selects data and losses to accelerate KD or enhance model robustness. In contrast, we select data for conducting different ways of distillation in our proposed method.

\paragraph{Continual Learning.} Continual learning (CL) for neural machine translation (NMT) aims at learning knowledge of new domains~\cite{thompson-etal-2019-overcoming,Liang_Zhao_Wang_Qiu_Li_2021_AAAI,cao2021continual} or languages~\cite{neubig-hu-2018-rapid,garcia-etal-2021-towards,huang-etal-2022-continual} without forgetting old knowledge. Our scenario also requires learning new knowledge but focuses on improving performance of the student on its test set instead.
Moreover, alleviating the negative impact of the less beneficial knowledge conveyed by ``weak'' teachers is essential in our scenario, which is hardly explored in CL for NMT. 
While our scenario is a multi-step process, multi-step CL is less explored in NMT~\cite{cao2021continual,Liang_Zhao_Wang_Qiu_Li_2021_AAAI}.
\citet{zeng-etal-2019-iterative} address a similar task of adapting from multiple out-of-domain models to a single in-domain model. Nevertheless, they assume the training data for the out-of-domain models are available, which is inaccessible in our scenario.
Besides, leveraging high-resource language NMT models to improve low-resource language translation has also attracted intensive efforts~\cite{neubig-hu-2018-rapid,lakew-etal-2019-adapting,liu-etal-2021-continual,huang-etal-2022-continual}, which can be a future extension of our method.

\section{Conclusion and Future Work}
To take advantage of increasingly available trained neural machine translation (NMT) models to improve one model of interest, we propose a novel method named continual knowledge distillation. Specially, knowledge from the trained models is transferred to the interested model via knowledge distillation in a sequential manner. Extensive experiments on two language pairs under homogeneous, heterogeneous, and malicious teacher settings show the effectiveness of our proposed method.

In the future, we will further explore the effect of the teacher model order. 
It is also worth involving more sophisticated  methods in knowledge filtration, such as gradient-based and meta-learning-based methods.

Moreover, it is also a promising research direction to exchange knowledge among all the models such that all of them achieve improvement.

\section*{Limitations}
There are some limitations that have yet to be addressed. Since we use the predicted probability distributions of the model output as a medium for \KANMT, the vocabulary of multiple models needs to be consistent. Overcoming it allows \KANMT to be extended to models with different language pairs and different modalities. Also, although our approach is robust to malicious models, there are more diverse and sophisticated attacks in 
real-world that require more research on defense. In addition, the teacher and student models must be trained on the same language pair. Further studies can consider more general scenarios without the above limitations. There are other approaches worth exploring in order to address the transfer of knowledge from models rather than their training data besides sequential manner. For example, it is also possible to explore various distillation methods like organizing teacher models into batches or pipelines.

\section*{Ethics Statement}
In practice, a provider may publicly release a model but may not wish its knowledge to be transferred into another one. Applying our method on such models will result in model stealing~\cite{he-2022-model-stealing} related ethical concerns. How to detect this kind of misconduct still needs further exploration.
Although sharing knowledge without exposing private data is one of the potential benefits of our method, models produced by our method are still vulnerable to attacks such as membership inference~\cite{hisamoto-etal-2020-membership}, and the private training data could still be stolen from the model.

\section*{Acknowledgments}
This work is supported by the National Key R\&D Program of China (2022ZD0160502) and the National Natural Science Foundation of China (No. 61925601, 62276152, 62236011). We thank all the reviewers for their valuable and insightful comments. We also thank Weihe Gao, Kaiyu Huang and Shuo Wang for their help.

\bibliography{custom}
\bibliographystyle{acl_natbib}

\clearpage
\appendix

\section{Datasets for Chinese-to-English Translation}
\label{sec:appendix:datasets}
The statistics of the datasets have been shown in Table~\ref{tab:corpora}.
The training data for the news, oral, internet, speech and subtitle domains are randomly sampled from LDC \footnote{LDC2002E18, LDC2003E07, LDC2003E14, part of LDC2004T07, LDC2004T08 and LDC2005T06}, AI Challenger 2018~\cite{aichallenger}, translation2019zh~\footnote{\url{https://github.com/brightmart/nlp_chinese_corpus}, version 1.0.}, TED transcripts~\cite{opus} and Subtitles~\cite{subtitles}, respectively. Newstest 2018 and NIST 02-09 from LDC are used as the development and test set for the news domain. The AI Challenger 2017 dataset is used as the test set for the oral domain. For the other corresponding domains, the development and test sets provided along with the training sets are used as development and test sets accordingly.

\section{Hyper-parameter Search}
\label{sec:appendix:hyper-parameter}

\begin{table}[!htbp]
  \centering
  \small
  \begin{threeparttable}
    \begin{tabular}{rr}
    \toprule
 $\mathbf{\alpha}$  & \multicolumn{1}{c}{\bf BLEU}
    \cr
    \midrule
    \midrule
    0.05 & 30.86
    \cr
    0.1 & \textbf{31.07} 
    \cr
    0.5 & 30.96
    \cr
    1  & 30.43
    \cr
    2  & 29.54
    \cr
    3 & 28.20
    \cr
    \bottomrule
    \end{tabular}
    \end{threeparttable}
      \captionof{table}{Results under different $\alpha$. The metrics are averaged over six configurations. }
      \label{tab:hyper}
\end{table}
\begin{table}[!htbp]
  \centering
  \small
  \begin{threeparttable}
    \begin{tabular}{lr}
    \toprule
 \multicolumn{1}{l}{$\mathbf{k_a : k_b}$}  &  \multicolumn{1}{c}{\bf BLEU}
    \cr
    \midrule
    \midrule
    {1 : 1} & \textbf{31.07} 
    \cr
    {1 : 0.5} & 31.03
    \cr
    {1 : 2} & 31.06
    \cr
    \bottomrule
    \end{tabular}
    \end{threeparttable}
      \captionof{table}{Results under different ratio of $k_a : k_b$. The metrics are averaged over six configurations. }
      \label{tab:hyper2}
\end{table}

For hyper-parameter $\alpha$ in Eq.~\ref{eqn:neg-kd}, we try multiple values in Table~\ref{tab:hyper}, and choose $0.1$ as the default value.

Since $\alpha$ regulates how much the student output distribution is pushed away from the negative distribution, we also try to regulate the proportion of positive and negative KD losses that work in a similar way. We modify Eq.~\ref{eqn:sum_all} to:
\begin{multline}
 \ell_{\mathrm{KF}}(\bm{\phi}_t, \bm{\theta}^{*}_t, D_{\mathrm{trans}}) 
=\\  k_a\ell_{\mathrm{KD}}(\bm{\phi}_t, \bm{\theta}^{*}_t, D^{+}_{\mathrm{trans}}) 
+  k_b\ell_{\mathrm{NEG}}(\bm{\phi}_t, \bm{\theta}^{*}_t, D^{-}_{\mathrm{trans}})
\end{multline}

By adjusting $k_a$ and $k_b$, we can regulate the weights of positive and negative losses. As shown in Table~\ref{tab:hyper2}, we still use the original settings since no significant performance improvement is found when adjusting $k_a : k_b$.

\section{Exploring Knowledge Filtration Quantification Function}
\label{appendixmetric}
\subsection{Examples}

\begin{table*}[!t]
  \centering
  \small
  \resizebox{0.93\textwidth}{!}{
  \begin{threeparttable}
  \label{tab:performance_comparison2}
    \begin{tabular}{l|lllllll}
    \toprule
    {\bf Source}  &\multicolumn{6}{l}{ \begin{CJK}{UTF8}{gbsn}每棵圣诞树上都挂满琳琅目的装点，但每棵树的顶端必定有一特大的星星  \end{CJK} }
    \\
    {\bf Target}  & \multicolumn{6}{l}{\it every christmas tree hung with dazzling  \colorbox{yellow}{decorations} \color{gray}{, but the top of each tree must have a tree big stars}}
    \\
    {\bf Teacher}   & {\bf Candidates:} & \colorbox{yellow}{decorations}$_{P=0.396}$ & ornaments$_{P=0.125}$ & costumes$_{P=0.033}$ & ar@@$_{P=0.032}$ & jewelry$_{P=0.022}$
    \\
    {\bf Student}    & {\bf Candidates:}& display$_{P=0.023}$ & car@@$_{P=0.0022}$ & displays$_{P=0.019}$ & 's$_{P=0.011}$ & \colorbox{yellow}{decorations}$_{P=0.010}$
    \\

    {\bf Loss}  & \multicolumn{4}{l}{$Q=0.396 \textgreater  Q=0.010 \Longrightarrow \langle y_j, \mathbf{y}_{<j}, \mathbf{x} \rangle$ belongs to $D^{+}_{\mathrm{trans}} \Longrightarrow \ell_{\mathrm{KD}}=2.542$  } & \multicolumn{2}{l}{// Teacher is informative.}
    \\
    \midrule \midrule
      {\bf Source}  &\multicolumn{6}{l}{ \begin{CJK}{UTF8}{gbsn}无量跌停并非没有前兆，前一交易日它的证券股价表现已经显得出奇地疲弱\end{CJK} }
    \\
    {\bf Target}  & \multicolumn{6}{l}{\it measureless limit is a  \colorbox{yellow}{precursor}\color{gray}{, in the previous session its securities share price performance appears ...}}
    \\
    {\bf Teacher}    & {\bf Candidates:}& fore@@$_{P=0.216}$ & sign$_{P=0.144}$ & single$_{P=0.104}$ & good@@$_{P=0.037}$ & major$_{P=0.029}$
    \\
    {\bf Student}    & {\bf Candidates:} & \colorbox{yellow}{pre@@}$_{P=0.533}$ & \colorbox{yellow}{precursor@@}$_{P=0.161}$ & aug@@$_{P=0.030}$ & omen$_{P=0.017}$ & pre$_{P=0.012}$
    \\

    {\bf Loss}  & 
    \multicolumn{4}{l}{
    $Q=0.003 \textless Q=0.161 \Longrightarrow \langle y_j, \mathbf{y}_{<j}, \mathbf{x} \rangle$ belongs to $D^{-}_{\mathrm{trans}} \Longrightarrow \ell_{\mathrm{NEG}}=\min\{0, \alpha-(-3.921)\}=0$
}  & \multicolumn{2}{l}{// Teacher is too unproductive, $\alpha=0.1$.}
    \\
    \midrule \midrule
       {\bf Source}  &\multicolumn{6}{l}{ \begin{CJK}{UTF8}{gbsn}这笔钱将提存立即中标人 \end{CJK} }
    \\
    {\bf Target}  & \multicolumn{6}{l}{\it money will be escrowed immediately to winning  \colorbox{yellow}{bidder}\color{gray}{.} }
    \\
    {\bf Teacher}    & {\bf Candidates:} & the@@$_{P=0.394}$ & mark$_{P=0.260}$ & be$_{P=0.034}$ & pay@@$_{P=0.034}$ & save$_{P=0.019}$
    \\
    {\bf Student}    & {\bf Candidates:} & target@@$_{P=0.331}$ & get@@$_{P=0.235}$ & sign@@$_{P=0.022}$ & put$_{P=0.021}$ & get$_{P=0.020}$
    \\

    {\bf Loss}  & \multicolumn{4}{l}{
    
    $Q=0.001 \textless Q=0.003 \Longrightarrow \langle y_j, \mathbf{y}_{<j}, \mathbf{x} \rangle$ belongs to $D^{-}_{\mathrm{trans}} \Longrightarrow \ell_{\mathrm{NEG}}=\min\{0, \alpha-0.085\}=\alpha-0.085$
    } & \multicolumn{2}{l}{// Teacher is somewhat informative, $\alpha=0.1$.}
    \\
    \bottomrule
    \end{tabular}
    \end{threeparttable}
  }
      \caption{ Three representative examples for illustrating the effectiveness of the knowledge filtration. Ground truth token $y_j$ and candidates matching with $y_j$ are highlighted in \colorbox{yellow}{yellow}. ``Student'' and ``Teacher'' show the top 5 predicted candidate tokens and their corresponding probabilities. $\alpha=0.1$ is the threshold here.}
      \label{tab:casestudy}
\end{table*}

\label{casestudy}
In Table~\ref{tab:casestudy}, we show three examples to demonstrate how 
the default quantification function (token-level cross entropy) works in knowledge filtration.
\begin{itemize}
  \item In the first case, we apply standard knowledge distillation because the teacher model assigns a higher probability of the ground truth token ``decorations'' than student, indicating a better distribution from the former.
  \item In the second case, the output from the teacher model is discarded because the negative KD loss exceeds the threshold. It might be a reasonable choice since the output of the teacher is too far from the ground truth token.
  \item In the third case, the teacher model have slightly worse predictions than students, motivating the student model not to make similar error-prone mistakes. 
\end{itemize}

\subsection{Alternatives to Quantification Function}
\begin{table*}[!t]
  \centering
  \small
  \resizebox{0.99\textwidth}{!}{
  \begin{threeparttable}
    \begin{tabular}{lllll}
    \toprule
 {\bf Metric}  & \makecell[l]{$\mathbbm{1}\{y_j=k \}$ \\ $\mathbbm{1}\{y_j=k^{\ast} \}$} & \makecell[l]{$\mathbbm{1}\{y_j \neq k \}$ \\ $\mathbbm{1}\{y_j \neq k^{\ast} \}$} & \makecell[l]{$\mathbbm{1}\{y_j=k \}$ \\ $\mathbbm{1}\{y_j \neq k^{\ast} \}$} & \makecell[l]{$\mathbbm{1}\{y_j \neq k \}$ \\ $\mathbbm{1}\{y_j=k^{\ast} \}$}
    \cr
    \midrule
    \midrule
    Trivial & + KD loss & + KD loss& + KD loss& + KD loss
    \cr
    Hard Label Matching & Discarded & Discarded & Discarded & + KD loss
    \cr
    \makecell[l]{Hard Label Matching (With Filtration)}  & Discarded & Discarded & - KD loss & + KD loss
    \cr
    Token-level CE & \makecell[l]{$\left\{ \begin{aligned} &\mbox{+ KD loss if } \Delta Q > 0 \\ &\mbox{Discarded if } \Delta Q \leq 0 \\ \end{aligned} \right.$} & \makecell[l]{$\left\{ \begin{aligned} &\mbox{+ KD loss if } \Delta Q > 0 \\ &\mbox{Discarded if } \Delta Q \leq 0 \\ \end{aligned} \right.$} & Discarded & + KD loss
    \cr
    \makecell[l]{Token-level CE (With Filtration)} & \makecell[l]{$\left\{ \begin{aligned} &\mbox{+KD loss if } \Delta Q > 0 \\ &\mbox{- KD loss if } \Delta Q \leq 0 \\ \end{aligned} \right.$} & \makecell[l]{$\left\{ \begin{aligned} &\mbox{+KD loss if } \Delta Q > 0 \\ &\mbox{- KD loss if } \Delta Q \leq 0 \\ \end{aligned} \right.$} & - KD loss & + KD loss
    \cr
    Hybrid Metric 1 & \makecell[l]{$\left\{ \begin{aligned} &\mbox{+KD loss if } \Delta Q > 0 \\ &\mbox{- KD loss if } \Delta Q \leq 0 \\ \end{aligned} \right.$} & \makecell[l]{$\left\{ \begin{aligned} &\mbox{+KD loss if } \Delta Q > 0 \\ &\mbox{Discarded if } \Delta Q \leq 0 \\ \end{aligned} \right.$} & - KD loss & + KD loss
    \cr
    Hybrid Metric 2 & \makecell[l]{$\left\{ \begin{aligned} &\mbox{+KD loss if } \Delta Q > 0 \\ &\mbox{Discarded if } \Delta Q \leq 0 \\ \end{aligned} \right.$} & \makecell[l]{$\left\{ \begin{aligned} &\mbox{+KD loss if } \Delta Q > 0 \\ &\mbox{- KD loss if } \Delta Q \leq 0 \\ \end{aligned} \right.$} & - KD loss & + KD loss
    \cr
    Hybrid Metric 3 & \makecell[l]{$\left\{ \begin{aligned} &\mbox{Discarded if } \Delta Q > 0 \\ &\mbox{- KD loss if } \Delta Q \leq 0 \\ \end{aligned} \right.$} & \makecell[l]{$\left\{ \begin{aligned} &\mbox{+KD loss if } \Delta Q > 0 \\ &\mbox{- KD loss if } \Delta Q \leq 0 \\ \end{aligned} \right.$} & - KD loss & + KD loss
    \cr
    \bottomrule
    \end{tabular}
    \end{threeparttable}
    }
    \caption{ Different metrics have different behaviors depending on the correctness of student's and teacher's prediction $k$ and $k^{\ast}$ for a given token $y_j$. ``+ KD loss'' and ``- KD loss'' mean positive and negative KD loss. ``$\Delta Q$'' denotes the difference in metric $f$ between the student and teacher model. $\mathbbm{1}$ is an indicator function.}
    \label{tab:kdm}
\end{table*}

The advantage of token-level cross entropy is that the predictions corresponding to the tokens in the transfer set $D_{\mathrm{trans}}$ can be divided into two mutually disjoint parts depending on the comparison between the predictions of teacher and student models. In contrast, hard label matching divides $D_{\mathrm{trans}}$ according to whether the teacher and student models predict the ground-truth token correctly, which will result in four parts due ties as shown in Table~\ref{tab:kdm}. 

Are the advantages of these two metrics beneficial for our task?  Is it possible to combine the beneficial properties? To answer the questions, we define several metrics in Table~\ref{tab:kdm} to compare these two metrics at a fine-grained level.  And the effects of these metrics are shown in Table~\ref{tab:kdmresult}. It could be found that token-level cross entropy always performs better because fewer samples are discarded such that more knowledge is transferred in knowledge distillation.

\begin{table}[!t]
  \centering
  \small
  \begin{threeparttable}
    \begin{tabular}{lr}
    \toprule
{\bf Metric}  & \multicolumn{1}{c}{\bf BLEU}
    \cr
    \midrule
    \midrule
    Trivial & 27.36
    \cr
    Hard Label Matching & 29.13
    \cr
    \makecell[l]{Hard Label  Matching (With Filtration)} & 30.37
    \cr
    Token-level CE & 30.69
    \cr
    \makecell[l]{Token-level CE  (With Filtration)} & \textbf{31.07} 
    \cr
    Hybrid Metric 1 & 31.05
    \cr
    Hybrid Metric 2 & 30.89
    \cr
    Hybrid Metric 3 & 30.22
    \cr
    \bottomrule
    \end{tabular}
    \end{threeparttable}
      \captionof{table}{Results for different metrics. The metrics are averaged over six configurations. }
      \vspace{-1em}
      \label{tab:kdmresult}
\end{table}

\section{Detailed Results}

\subsection{Chinese-to-English Translation}
\label{full}

\begin{table*}[!h]
    \centering
    \resizebox{0.99\textwidth}{!}{
    \begin{threeparttable}
      \small\setlength\tabcolsep{3.5pt} 
      \begin{tabular}{cl|lrlrlrlrlrlr|lr} 
      \toprule
      \multirow{2}[3]{*}{\bf Step}&\multirow{2}[3]{*}{\bf Method}&
      \multicolumn{2}{c}{\bf BCDE$\rightarrow$A}&\multicolumn{2}{c}{\bf CBDE$\rightarrow$A}&\multicolumn{2}{c}{\bf ACDE$\rightarrow$B}&\multicolumn{2}{c}{\bf CADE$\rightarrow$B}&\multicolumn{2}{c}{\bf ABDE$\rightarrow$C}&\multicolumn{2}{c|}{\bf BADE$\rightarrow$C}&\multicolumn{2}{c}{\bf Average}\\
      \cmidrule(lr){3-4} \cmidrule(lr){5-6} \cmidrule(lr){7-8}
      \cmidrule(lr){9-10}
      \cmidrule(lr){11-12}
      \cmidrule(lr){13-14}
      \cmidrule(lr){15-16}
      && \multicolumn{1}{c}{{\bf BLEU$\mathbf{\uparrow}$}} & \multicolumn{1}{c}{{{\bf AD$\mathbf{\downarrow}$}}} & \multicolumn{1}{c}{{\bf BLEU$\mathbf{\uparrow}$}} & \multicolumn{1}{c}{{{\bf AD$\mathbf{\downarrow}$}}} & \multicolumn{1}{c}{{\bf BLEU$\mathbf{\uparrow}$}} & \multicolumn{1}{c}{{{\bf AD$\mathbf{\downarrow}$}}} & \multicolumn{1}{c}{{\bf BLEU$\mathbf{\uparrow}$}} & \multicolumn{1}{c}{{{\bf AD$\mathbf{\downarrow}$}}} & \multicolumn{1}{c}{{\bf BLEU$\mathbf{\uparrow}$}} & \multicolumn{1}{c}{{{\bf AD$\mathbf{\downarrow}$}}} & \multicolumn{1}{c}{{\bf BLEU$\mathbf{\uparrow}$}} & \multicolumn{1}{c|}{{{\bf AD$\mathbf{\downarrow}$}}} & \multicolumn{1}{c}{{\bf BLEU$\mathbf{\uparrow}$}} & \multicolumn{1}{c}{{{\bf AD$\mathbf{\downarrow}$}}}\\
      \midrule\midrule
      0 & & 42.84 & & 42.84 & & 27.53 & & 27.53 & & 18.06 & & 18.06 & & 29.48 & /
      \cr
      \midrule

      \multirow{4}{*}{1} & KD & 46.19$_{\text{3.35}}$ & 0.00 & 44.47$_{\text{1.63}}$ & 0.00 & 24.32$_{\text{-3.21}}$ & 3.21 & 26.17$_{\text{-1.36}}$ & 1.36 & 17.06$_{\text{-1.00}}$ & 1.00 & 19.22$_{\text{1.16}}$ & 0.00 & 29.57$_{\text{0.09}}$ & 0.93

      \cr
      & EWC & 46.09$_{\text{3.25}}$ & 0.00 & 44.59$_{\text{1.75}}$ & 0.00 & 24.32$_{\text{-3.21}}$ & 3.21 & 26.26$_{\text{-1.27}}$ & 1.27 & 17.12$_{\text{-0.94}}$ & 0.94 & 19.35$_{\text{1.29}}$ & 0.00 & 29.57$_{\text{0.15}}$ & 0.90
      \cr
      & CL-NMT & 46.14$_{\text{3.30}}$ & 0.00 & 44.53$_{\text{1.69}}$ & 0.00 & 24.28$_{\text{-3.25}}$ & 3.25 & 26.21$_{\text{-1.32}}$ & 1.32 & 17.09$_{\text{-0.97}}$ & 0.97 & 19.17$_{\text{1.11}}$ & 0.00 & 29.62$_{\text{0.09}}$ & 0.92
      
      \cr
      & Ours & 46.00$_{\text{3.16}}$ & 0.00 & 45.88$_{\text{3.04}}$ & 0.00 & 28.01$_{\text{0.48}}$ & 0.00 & 28.17$_{\text{0.64}}$ & 0.00 & 18.98$_{\text{0.92}}$ & 0.00 & 19.36$_{\text{1.30}}$ & 0.00 & 31.07$_{\text{1.59}}$ & 0.00
      \cr
      \midrule
      \multirow{4}{*}{2} & KD & 44.62$_{\text{1.78}}$ & 1.57 & 46.28$_{\text{3.44}}$ & 0.00 & 26.11$_{\text{-1.42}}$ & 3.21 & 23.96$_{\text{-3.57}}$ & 3.57 & 19.33$_{\text{1.27}}$ & 1.00 & 17.25$_{\text{-0.81}}$ & 1.97 & 29.59$_{\text{0.11}}$ & 1.89
      \cr
      & EWC & 45.80$_{\text{2.96}}$ & 0.29 & 46.26$_{\text{3.42}}$ & 0.00 & 25.28$_{\text{-2.25}}$ & 3.21 & 25.92$_{\text{-1.61}}$ & 1.61 & 18.13$_{\text{0.07}}$ & 0.94 & 18.52$_{\text{0.46}}$ & 0.83 & 30.65$_{\text{0.51}}$ & 1.15
      \cr
      & CL-NMT & 45.24$_{\text{2.40}}$ & 0.90 & 45.48$_{\text{2.64}}$ & 0.00 & 27.67$_{\text{0.14}}$ & 3.25 & 27.70$_{\text{0.17}}$ & 1.32 & 19.09$_{\text{1.03}}$ & 0.97 &  18.69$_{\text{0.63}}$ & 0.48 & 29.99$_{\text{1.17}}$ & 1.15
      
      \cr
      & Ours & 45.89$_{\text{3.05}}$ & 0.11 & 46.08$_{\text{3.24}}$ & 0.00 & 28.28$_{\text{0.75}}$ & 0.00 & 28.50$_{\text{0.97}}$ & 0.00 & 19.18$_{\text{1.12}}$ & 0.00 & 19.36$_{\text{1.30}}$ & 0.00 & 31.22$_{\text{1.74}}$ & 0.02
      \cr
      \midrule
      \multirow{4}{*}{3} & KD & 39.16$_{\text{-3.68}}$ & 7.03 & 39.11$_{\text{-3.73}}$ & 7.17 & 21.76$_{\text{-5.77}}$ & 7.56 & 21.69$_{\text{-5.84}}$ & 5.84  & 16.14$_{\text{-1.92}}$ & 4.19 & 16.12$_{\text{-1.94}}$ & 3.10 & 25.66$_{\text{-3.81}}$ & 5.82
      \cr
      & EWC & 43.88$_{\text{1.04}}$ & 2.21 & 44.02$_{\text{1.18}}$ & 2.24 & 24.48$_{\text{-3.05}}$ & 4.01 & 24.27$_{\text{-3.26}}$ & 3.26 & 17.76$_{\text{-0.30}}$ & 1.31 & 18.09$_{\text{0.03}}$ & 1.26 & 29.91$_{\text{-0.73}}$ & 2.38
      \cr
      
      & CL-NMT & 43.91$_{\text{1.07}}$ & 2.23 & 43.83$_{\text{0.99}}$ & 1.65 & 27.23$_{\text{-0.3}}$ & 3.69 & 27.32$_{\text{-0.21}}$ & 1.70 & 18.45$_{\text{0.39}}$ & 1.61 & 18.71$_{\text{0.65}}$ & 0.48 & 28.75$_{\text{0.43}}$ & 1.89
      
      \cr
      & Ours & 45.89$_{\text{3.05}}$ & 0.11 & 46.08$_{\text{3.24}}$ & 0.00 & 28.41$_{\text{0.88}}$ & 0.00 & 28.48$_{\text{0.95}}$ & 0.02 & 19.15$_{\text{1.09}}$ & 0.03 & 18.98$_{\text{0.92}}$ & 0.38 & 31.17$_{\text{1.69}}$ & 0.09
      \cr
      \midrule
      \multirow{4}{*}{4} & KD & 30.57$_{\text{-12.27}}$ & 15.62 & 30.31$_{\text{-12.53}}$ & 15.97 & 22.71$_{\text{-4.82}}$ & 7.56 & 22.66$_{\text{-4.87}}$ & 5.84 & 13.88$_{\text{-4.18}}$ & 6.45 & 13.99$_{\text{-4.07}}$ & 5.23 & 22.35$_{\text{-7.12}}$ & 9.45
      \cr
      & EWC & 41.13$_{\text{-1.71}}$ & 4.96 & 37.41$_{\text{-5.43}}$ & 8.85 & 24.89$_{\text{-2.64}}$ & 4.01 & 24.96$_{\text{-2.57}}$ & 3.26 & 17.24$_{\text{-0.82}}$ & 1.83 & 17.59$_{\text{-0.47}}$ & 1.76 & 29.85$_{\text{-2.27}}$ & 4.11
      \cr
      & CL-NMT & 43.13$_{\text{0.29}}$ & 3.01 & 42.99$_{\text{0.15}}$ & 2.49 & 27.90$_{\text{0.37}}$ & 3.69 & 27.93$_{\text{0.4}}$ & 1.70 & 18.59$_{\text{0.53}}$ & 1.61 & 18.58$_{\text{0.52}}$ & 0.61 & 27.20$_{\text{0.38}}$ & 2.19
      
      \cr
      & Ours & \textbf{45.89}$_{\text{3.05}}$ & \textbf{0.11} & \textbf{46.08}$_{\text{3.24}}$ & \textbf{0.00} & \textbf{28.49}$_{\text{0.96}}$ & \textbf{0.00} & \textbf{28.51}$_{\text{0.98}}$ & \textbf{0.02} & \textbf{19.15}$_{\text{1.09}}$ & \textbf{0.03} & \textbf{18.98}$_{\text{0.92}}$ & \textbf{0.38} & \textbf{31.18}$_{\text{1.71}}$ & \textbf{0.09}
      \cr
      \bottomrule
      \end{tabular}
      \end{threeparttable}
      }
      \caption{Results of \textit{Chinese-to-English} translation under \textit{homogeneous} teacher setting. ``BCDE$\rightarrow$A'' denotes $A$ is the student model and $B$, $C$, $D$, and $E$ are teacher models in step $1$ to $4$,  respectively. ``$\Delta$'' denotes $\Delta$BLEU compared with step $0$ (i.e., initial student model), and $\Delta$BLEU scores are also reported as subscript numbers. ``AD'' is the accumulative degradation defined in Eq.~\ref{eqn:AQD}, which is the lower the better. The last two columns are numbers averaged row-wise. Best results in step $4$ are in {\bf bold}.}
      \label{tab:main2_total}
\end{table*}

The full results for the Chinese-to-English homogeneous model setting for all configurations are shown in Table~\ref{tab:main2_total}.
Our experiments are conducted on NVIDIA
A100 GPUs. Each distillation process requires 48 GPU hours and is run only once due to computational budgets.

\subsection{Larger Scale Chinese-to-English Translation}
\label{app:larger}

\begin{table*}[!t]
    \centering
    \resizebox{0.99\textwidth}{!}{
    \begin{threeparttable}
      \small\setlength\tabcolsep{3.5pt} 
      \begin{tabular}{cl|lrlrlrlrlrlr|lr} 
      \toprule
      \multirow{2}[3]{*}{\bf Step}&\multirow{2}[3]{*}{\bf Method}&
      \multicolumn{2}{c}{\bf GHIJ$\rightarrow$F}&\multicolumn{2}{c}{\bf HGIJ$\rightarrow$F}&\multicolumn{2}{c}{\bf FHIJ$\rightarrow$G}&\multicolumn{2}{c}{\bf HFIJ$\rightarrow$G}&\multicolumn{2}{c}{\bf FGIJ$\rightarrow$H}&\multicolumn{2}{c|}{\bf GFIJ$\rightarrow$H}&\multicolumn{2}{c}{\bf Average}\\
      \cmidrule(lr){3-4} \cmidrule(lr){5-6} \cmidrule(lr){7-8}
      \cmidrule(lr){9-10}
      \cmidrule(lr){11-12}
      \cmidrule(lr){13-14}
      \cmidrule(lr){15-16}
      && \multicolumn{1}{c}{{\bf BLEU$\mathbf{\uparrow}$}} & \multicolumn{1}{c}{{{\bf AD$\mathbf{\downarrow}$}}} & \multicolumn{1}{c}{{\bf BLEU$\mathbf{\uparrow}$}} & \multicolumn{1}{c}{{{\bf AD$\mathbf{\downarrow}$}}} & \multicolumn{1}{c}{{\bf BLEU$\mathbf{\uparrow}$}} & \multicolumn{1}{c}{{{\bf AD$\mathbf{\downarrow}$}}} & \multicolumn{1}{c}{{\bf BLEU$\mathbf{\uparrow}$}} & \multicolumn{1}{c}{{{\bf AD$\mathbf{\downarrow}$}}} & \multicolumn{1}{c}{{\bf BLEU$\mathbf{\uparrow}$}} & \multicolumn{1}{c}{{{\bf AD$\mathbf{\downarrow}$}}} & \multicolumn{1}{c}{{\bf BLEU$\mathbf{\uparrow}$}} & \multicolumn{1}{c|}{{{\bf AD$\mathbf{\downarrow}$}}} & \multicolumn{1}{c}{{\bf BLEU$\mathbf{\uparrow}$}} & \multicolumn{1}{c}{{{\bf AD$\mathbf{\downarrow}$}}}\\
      \midrule\midrule
      0 & & 43.16 & & 43.16 & & 31.35 & & 31.35 & & 23.8 & & 23.8 & & 32.77 & /
      \cr
      \midrule

      \multirow{4}{*}{1} & KD & 43.90$_{\text{0.74}}$ & 0.00 & 41.54$_{\text{-1.62}}$ & 1.62 & 29.92$_{\text{-1.43}}$ & 1.43 & 29.70$_{\text{-1.65}}$ & 1.65 & 23.70$_{\text{-0.10}}$ & 0.10 & 23.69$_{\text{-0.11}}$ & 0.11 & 32.08$_{\text{-0.70}}$ & 0.82

\cr
      & EWC & 43.99$_{\text{0.83}}$ & 0.00 & 41.65$_{\text{-1.51}}$ & 1.51 & 29.90$_{\text{-1.45}}$ & 1.45 & 29.73$_{\text{-1.62}}$ & 1.62 & 23.70$_{\text{-0.10}}$ & 0.10 & 23.66$_{\text{-0.14}}$ & 0.14 & 32.17$_{\text{-0.66}}$ & 0.80
      \cr
      & CL-NMT & 44.08$_{\text{0.92}}$ & 0.00 & 41.75$_{\text{-1.41}}$ & 1.41 & 29.92$_{\text{-1.43}}$ & 1.43 & 29.77$_{\text{-1.58}}$ & 1.58 & 23.70$_{\text{-0.10}}$ & 0.10 & 23.76$_{\text{-0.04}}$ & 0.04 & 32.11$_{\text{-0.60}}$ & 0.76
      
      \cr
      & Ours & 44.25$_{\text{1.09}}$ & 0.00 & 44.05$_{\text{0.89}}$ & 0.00 & 31.51$_{\text{0.16}}$ & 0.00 & 31.35$_{\text{0.00}}$ & 0.00 & 23.81$_{\text{0.01}}$ & 0.00 & 23.77$_{\text{-0.03}}$ &0.03 & 33.12$_{\text{0.35}}$ & 0.01
      \cr
      \midrule
      \multirow{4}{*}{2} & KD & 42.19$_{\text{-0.97}}$ & 1.71 & 44.06$_{\text{0.90}}$ & 1.62 & 28.91$_{\text{-2.44}}$ & 2.44 & 28.94$_{\text{-2.41}}$ & 2.41 & 22.86$_{\text{-0.94}}$ & 0.94 & 22.84$_{\text{-0.96}}$ & 0.96 & 31.63$_{\text{-1.14}}$ & 1.68
      \cr
      & EWC & 43.16$_{\text{0.00}}$ & 0.83 & 43.03$_{\text{-0.13}}$ & 1.51 & 27.83$_{\text{-3.52}}$ & 3.52 &30.21$_{\text{-1.14}}$ & 1.62 & 23.01$_{\text{-0.79}}$ & 0.79 & 23.71$_{\text{-0.09}}$ & 0.14 & 32.40$_{\text{-0.94}}$ & 1.40
      \cr
      & CL-NMT & 44.05$_{\text{0.89}}$ & 0.04 & 44.09$_{\text{0.93}}$ & 1.41 & 29.46$_{\text{-1.89}}$ & 1.89 & 29.64$_{\text{-1.71}}$ & 1.71 & 23.58$_{\text{-0.22}}$ & 0.22 &  23.60$_{\text{-0.20}}$ & 0.20 & 31.83$_{\text{-0.37}}$ & 0.91
      
      \cr
      & Ours & 44.45$_{\text{1.29}}$ & 0.00 & 44.44$_{\text{1.28}}$ & 0.00 & 31.51$_{\text{0.16}}$ & 0.00 & 31.44$_{\text{0.09}}$ & 0.00 & 24.14$_{\text{0.34}}$ & 0.00 & 24.11$_{\text{0.31}}$ & 0.03 & 33.35$_{\text{0.58}}$ & 0.01
      \cr
      \midrule
      \multirow{4}{*}{3} & KD & 35.13$_{\text{-8.03}}$ & 8.77 & 34.86$_{\text{-8.30}}$ & 10.82 & 22.82$_{\text{-8.53}}$ & 8.53 & 22.78$_{\text{-8.57}}$ & 8.57  & 18.30$_{\text{-5.50}}$ & 5.50 & 18.28$_{\text{-5.52}}$ & 5.52 & 25.36$_{\text{-7.41}}$ & 7.95
      \cr
      & EWC & 41.58$_{\text{-1.58}}$ & 2.42 & 40.93$_{\text{-2.23}}$ & 3.61 & 27.03$_{\text{-4.32}}$ & 4.32 & 29.28$_{\text{-2.07}}$ & 2.55 & 22.05$_{\text{-1.75}}$ & 1.75 & 23.16$_{\text{-0.64}}$ & 0.69 & 29.89$_{\text{-2.10}}$ & 2.56
      \cr
      & CL-NMT & 41.30$_{\text{-1.86}}$ & 2.79 & 43.38$_{\text{0.22}}$ & 2.11 & 29.47$_{\text{-1.88}}$ & 1.89 & 24.25$_{\text{-7.10}}$ & 7.10 & 22.68$_{\text{-1.12}}$ & 1.12 & 18.25$_{\text{-5.55}}$ & 5.55 & 30.67$_{\text{-2.88}}$ & 3.42
      
      \cr
      & Ours & 44.63$_{\text{1.47}}$ & 0.00 & 44.47$_{\text{1.31}}$ & 0.00 & 31.48$_{\text{0.13}}$ & 0.03 & 31.50$_{\text{0.15}}$ & 0.00 & 24.04$_{\text{0.24}}$ & 0.10 & 24.12$_{\text{0.32}}$ & 0.03 & 33.37$_{\text{0.60}}$ & 0.03
      \cr
      \midrule
      \multirow{4}{*}{4} & KD & 39.45$_{\text{-3.71}}$ & 8.77 & 39.12$_{\text{-4.04}}$ & 10.82 & 26.67$_{\text{-4.68}}$ & 8.53 & 26.79$_{\text{-4.56}}$ & 8.57 & 20.39$_{\text{-3.41}}$ & 5.50 & 20.33$_{\text{-3.47}}$ & 5.52 & 28.79$_{\text{-3.98}}$ & 7.95
      \cr
      & EWC & 41.99$_{\text{-1.17}}$ & 2.42 & 41.40$_{\text{-1.76}}$ & 3.61 & 29.66$_{\text{-1.69}}$ & 4.32 & 30.77$_{\text{-0.58}}$ & 2.55 & 22.04$_{\text{-1.76}}$ & 1.76 & 23.37$_{\text{-0.43}}$ & 0.69 & 31.01$_{\text{-1.23}}$ & 2.56
      \cr
      & CL-NMT & 42.69$_{\text{-0.47}}$ & 2.79 & 43.60$_{\text{0.44}}$ & 2.11 & 30.96$_{\text{-0.39}}$ & 1.89 & 27.08$_{\text{-4.27}}$ & 7.10 & 22.99$_{\text{-0.81}}$ & 1.12 & 18.75$_{\text{-5.05}}$ & 5.55 & 31.54$_{\text{-1.76}}$ & 3.42
      
      \cr
      & Ours & \textbf{44.65}$_{\text{1.49}}$ & \textbf{0.00} & \textbf{44.54}$_{\text{1.38}}$ & \textbf{0.00} & \textbf{31.53}$_{\text{0.18}}$ & \textbf{0.03} & \textbf{31.55}$_{\text{0.20}}$ & \textbf{0.00} & \textbf{24.15}$_{\text{0.35}}$ & \textbf{0.10} & \textbf{24.14}$_{\text{0.34}}$ & \textbf{0.03} & 33.43$_{\text{0.66}}$  & \textbf{0.03}
      \cr
      \bottomrule
      \end{tabular}
      \end{threeparttable}
      }
      \caption{Results on larger Chinese-to-English datasets. ``GHIJ$\rightarrow$F'' denotes $F$ is the student model and model $G$ to $J$ are teacher models in step 1 to 4, respectively.}
      \label{tab:detail_big}
\end{table*}

\begin{table}[!t]
  \centering
  \small
  \resizebox{0.48\textwidth}{!}{
      \begin{threeparttable}
      \begin{tabular}{clrrr}
      \toprule
      \textbf{Model} & \textbf{Domain} & \textbf{Training} & \textbf{Dev.} &
      \textbf{Test}  \\
      \midrule
      $F$ &  News & 20,000,000  & 4,000 & 13,000
       \\
      $G$ &  Oral & 12,500,000 & 4,000 & 12,000
      \\
      $H$ &  Internet & 5,200,000 & 4,000& 13,000
      \\ 
      $I$ &  Speech & 220,000 & 4,000  & 5,000
      \\
      $J$ &  Subtitle & 300,000 & 4,000 & 4,000  \\
      \bottomrule
      \end{tabular}
      \end{threeparttable}}

      \caption{The domain, training and evaluation corpora of Chinese-to-English translation large-scale experiments. Compare with Table~\ref{tab:corpora}, news, oral and Internet datasets are replaced with larger datasets to test the scalability. Other settings are kept identical.}
      \label{tab:bigger}
  \end{table}

\begin{table*}[!t]
  \centering
  \resizebox{0.99\textwidth}{!}{
  \begin{threeparttable}
    \small\setlength\tabcolsep{3.5pt} 
    \begin{tabular}{cl|lrlrlrlrlrlr|lr} 
    \toprule
    \multirow{2}[3]{*}{\bf Step}&\multirow{2}[3]{*}{\bf Method}&
    \multicolumn{2}{c}{\bf LMN$\rightarrow$K}&\multicolumn{2}{c}{\bf MLN$\rightarrow$K}&\multicolumn{2}{c}{\bf KMN$\rightarrow$L}&\multicolumn{2}{c}{\bf MKN$\rightarrow$L}&\multicolumn{2}{c}{\bf KLN$\rightarrow$M}&\multicolumn{2}{c|}{\bf LKN$\rightarrow$M}&\multicolumn{2}{c}{\bf Average}\\
    \cmidrule(lr){3-4} \cmidrule(lr){5-6} \cmidrule(lr){7-8}
    \cmidrule(lr){9-10}
    \cmidrule(lr){11-12}
    \cmidrule(lr){13-14}
    \cmidrule(lr){15-16}
    && \multicolumn{1}{c}{{\bf BLEU$\mathbf{\uparrow}$}} & \multicolumn{1}{c}{{{\bf AD$\mathbf{\downarrow}$}}} & \multicolumn{1}{c}{{\bf BLEU$\mathbf{\uparrow}$}} & \multicolumn{1}{c}{{{\bf AD$\mathbf{\downarrow}$}}} & \multicolumn{1}{c}{{\bf BLEU$\mathbf{\uparrow}$}} & \multicolumn{1}{c}{{{\bf AD$\mathbf{\downarrow}$}}} & \multicolumn{1}{c}{{\bf BLEU$\mathbf{\uparrow}$}} & \multicolumn{1}{c}{{{\bf AD$\mathbf{\downarrow}$}}} & \multicolumn{1}{c}{{\bf BLEU$\mathbf{\uparrow}$}} & \multicolumn{1}{c}{{{\bf AD$\mathbf{\downarrow}$}}} & \multicolumn{1}{c}{{\bf BLEU$\mathbf{\uparrow}$}} & \multicolumn{1}{c|}{{{\bf AD$\mathbf{\downarrow}$}}} & \multicolumn{1}{c}{{\bf BLEU$\mathbf{\uparrow}$}} & \multicolumn{1}{c}{{{\bf AD$\mathbf{\downarrow}$}}}
    \\
    \midrule\midrule
    0 & & 31.10 & & 31.10 & & 30.04 & & 30.04 & & 30.72 & & 30.72 & & 30.62 & /
    \cr
    \midrule

    \multirow{4}{*}{1} & KD & 31.18$_{\text{0.08}}$ & 0.00 & 31.57$_{\text{0.47}}$ & 0.00 & 29.62$_{\text{-0.42}}$ & 0.42 & 29.84$_{\text{-0.20}}$ & 0.20 & 31.33$_{\text{0.61}}$ & 0.00 & 30.53$_{\text{-0.19}}$ & 0.19 & 30.68$_{\text{0.06}}$ & 0.13
\cr
    & EWC & 31.34$_{\text{0.24}}$ & 0.00 & 31.58$_{\text{0.48}}$ & 0.00 & 29.61$_{\text{-0.43}}$ & 0.43 & 29.85$_{\text{-0.19}}$ & 0.19 & 31.33$_{\text{0.61}}$ & 0.00 & 30.24$_{\text{-0.48}}$ &0.48 & 30.85$_{\text{0.04}}$ & 0.18

    \cr
    & CL-NMT & 31.50$_{\text{0.40}}$ & 0.00 & 31.59$_{\text{0.49}}$ & 0.00 & 29.62$_{\text{-0.42}}$ & 0.42 & 29.87$_{\text{-0.17}}$ & 0.17 & 31.33$_{\text{0.61}}$ & 0.00 & 31.18$_{\text{0.46}}$ & 0.00 & 30.66$_{\text{0.23}}$ & 0.10
   
    \cr
    & Ours & 31.78$_{\text{0.68}}$ & 0.00 & 31.82$_{\text{0.72}}$ & 0.00 & 30.36$_{\text{0.32}}$ & 0.00 & 30.43$_{\text{0.39}}$ & 0.00 & 31.38$_{\text{0.66}}$ & 0.00 & 31.34$_{\text{0.62}}$ & 0.00 & 31.19$_{\text{0.57}}$ & 0.00
    \cr
    \midrule
    \multirow{4}{*}{2} & KD & 30.96 $_{\text{-0.14}}$ & 0.22 & 30.76$_{\text{-0.34}}$ & 0.81 & 30.22$_{\text{0.18}}$ & 0.42 & 30.34$_{\text{0.30}}$ & 0.20 & 30.54$_{\text{-0.18}}$ & 0.79 & 31.30$_{\text{0.58}}$ & 0.19 & 30.69$_{\text{0.07}}$ & 0.44
    \cr
    & EWC & 31.26$_{\text{0.16}}$ & 0.08 & 31.26$_{\text{0.16}}$ & 0.32 & 30.34$_{\text{0.30}}$ & 0.43 &29.78$_{\text{-0.26}}$ & 0.26 & 30.64$_{\text{-0.08}}$ & 0.69 & 30.42$_{\text{-0.30}}$ & 0.48 & 30.79$_{\text{0.00}}$ & 0.37
    \cr
    & CL-NMT & 31.55$_{\text{0.45}}$ & 0.00 & 30.86$_{\text{-0.24}}$ & 0.73 & 29.95$_{\text{-0.09}}$ & 0.42 & 30.08$_{\text{0.04}}$ & 0.17 & 31.02$_{\text{0.30}}$ & 0.31 & 31.30$_{\text{0.58}}$ & 0.00 & 30.62$_{\text{0.17}}$ & 0.27
    
    \cr
    & Ours & 31.87$_{\text{0.77}}$ & 0.00 & 31.93$_{\text{0.83}}$ & 0.00 & 30.46$_{\text{0.42}}$ & 0.00 & 30.54$_{\text{0.50}}$ & 0.00 &31.38$_{\text{0.66}}$ & 0.00 & 31.33$_{\text{0.61}}$ & 0.01 & 31.25$_{\text{0.63}}$ & 0.00
    \cr
    \midrule
    \multirow{4}{*}{3} & KD & 23.74$_{\text{-7.36}}$ & 7.44 & 23.64$_{\text{-7.46}}$ & 7.93 & 23.11$_{\text{-6.93}}$ & 7.53 & 24.62$_{\text{-5.42}}$ & 5.92  & 26.00$_{\text{-4.72}}$ & 5.33 & 27.81$_{\text{-2.92}}$ & 3.68 & 24.82$_{\text{-5.80}}$ & 6.31
    \cr
    & EWC & 29.52$_{\text{-1.58}}$ & 1.82 & 29.63$_{\text{-1.47}}$ & 1.95 & 29.84$_{\text{-0.20}}$ & 0.92 & 28.88$_{\text{-1.16}}$ & 1.16 & 29.88$_{\text{-0.84}}$ & 1.45 & 30.10$_{\text{-0.62}}$ & 0.80 & 28.77$_{\text{-0.98}}$ & 1.35
    \cr
    & CL-NMT & 28.64$_{\text{-2.46}}$ & 2.91 & 30.30$_{\text{-0.80}}$ & 1.29 & 30.00$_{\text{-0.04}}$ & 0.42 & 25.07$_{\text{-4.97}}$ & 5.18 & 31.15$_{\text{0.43}}$ & 0.31 & 27.47$_{\text{-3.25}}$ & 3.83 & 29.64$_{\text{-1.85}}$ & 2.32
    
    \cr
    & Ours & \textbf{31.89}$_{\text{0.79}}$ & \textbf{0.00} & \textbf{31.94}$_{\text{0.84}}$ & \textbf{0.00} & \textbf{30.47}$_{\text{0.43}}$ & \textbf{0.00} & \textbf{30.55}$_{\text{0.51}}$ & \textbf{0.00} & \textbf{31.52}$_{\text{0.80}}$ & \textbf{0.00} & \textbf{31.54}$_{\text{0.82}}$ & \textbf{0.01} & \textbf{31.32}$_{\text{0.70}}$ & \textbf{0.00}
    \cr
    \bottomrule
    \end{tabular}
    \end{threeparttable}
    }
    \caption{Results on German-to-English datasets. ``LMN$\rightarrow$K'' denotes $K$ is the student model and model $L$, $M$ and $N$ are teacher models in step 1 to 3 respectively.}
    \label{tab:detail_ende}
\end{table*}

\begin{table}[!t]
\centering
\small
    \begin{threeparttable}
    \begin{tabular}{clrrr}
    \toprule
    \textbf{Model} & \textbf{Domain} & \textbf{Training} & \textbf{Dev.} &
    \textbf{Test}  \\
    \midrule
    $K$ &  News & 4,500,000  & 4,000 & 8,000
     \\
    $L$ &  Multiple & 4,300,000 & 4,000 & 8,000
    \\
    $M$ &  Europarl & 1,300,000 & 4,000& 8,000
    \\ 
    $N$ &  Tanzil & 500,000 & 4,000  & 8,000
    \\
    \bottomrule
    \end{tabular}
    \end{threeparttable}
    \captionof{table}{The domain, training and evaluation corpora of the four Transformer-base models used in the German-to-English translation experiments.
    Other settings are kept identical to Chinese-to-English translation. 
    }
    \label{tab:de}
\end{table}

To illustrate the scalability of our method, we repeat Chinese-to-English translation experiments under homogeneous model setting (Sec.~\ref{sec:zh-en-translation}) using larger and complete corpora without sampling. As shown in Table~\ref{tab:bigger}, we choose five full datasets for news, oral, internet, speech and subtitle domains, respectively: WMT20, AI Challenger 2018, translation2019zh, TED transcripts and Subtitles. Compared with Table~\ref{tab:corpora}, only the training set of model $F$,$G$ and $H$ are replaced with bigger datasets. And the validation sets, test sets and other settings are kept identical. $F$, $G$, $H$, $I$, $J$ are all Transformer-base models. The three models $F$, $G$ and $H$ are combined in different orders to form six groups of experiments. To simulate the most challenging scenario, $I$ and $J$ with weaker performance are added at the end of these experiments to test the performance of our method on poor models. According to the task definition illustrated above, the transfer set of each set of experiments is the training set of the according student model.

The results are shown in Table~\ref{tab:detail_big}. Similar to results in Sec.~\ref{sec:zh-en-translation}, our method outperforms baselines significantly and consistently, justifying that our method is scalable and generalizes to different corpus sizes. For example, without using any extra data, our method yield up to +1.49, +0.20, +0.35 BLEU scores on WMT20, AI Challenger 2018 and translation2019zh datasets, respectively.

\subsection{German-to-English Translation}
\label{app:ende}

As shown in Table~\ref{tab:de}, the training data for German-to-English translation are from WMT16, TildeMODEL v2018~\cite{tilde}, Tanzil v1 and Europarl~\cite{europarl}, respectively. We sample $4,000$ sentences from the original corpus as the development set and $8,000$ sentences as the test set. Model $N$ has weaker performance. Other settings are kept identical to the Chinese-to-English translation experiments. $K$, $L$, $M$, $N$ are all Transformer-base models. The three models $K$, $L$ and $M$ are combined in different orders to form six groups of experiments. To simulate the most challenging scenario, $N$ with weaker performance is added at the end of these experiments to test the performance of our method on poor models. According to the task definition illustrated above, the transfer set of each set of experiments is the training set of the according student model.

As shown in Table~\ref{tab:detail_ende}, our method performs similarly on the German-to-English language pair as it does on the Chinese-to-English language pair under the homogeneous model setting. We also repeat the experiments under the heterogeneous and malicious model settings. As shown in Table~\ref{tab:arc_ende} and Table~\ref{tab:malicious_ende}, our method is also superior to the baselines under these settings for German-to-English translation.

\begin{table}[!t]
\centering
\resizebox{0.48\textwidth}{!}{
\begin{threeparttable}
        \small\setlength\tabcolsep{3pt}
        \begin{tabular}{l|lrlrlr}
        \toprule
        \multirow{2}[3]{*}{\bf
        Method}&
        \multicolumn{2}{c}{\bf Transformer-base}&\multicolumn{2}{c}{\bf RNN}&\multicolumn{2}{c}{\bf Transformer-big} \\
        \cmidrule(lr){2-3} \cmidrule(lr){4-5} \cmidrule(lr){6-7}
        &$\mathbf{BLEU\uparrow}$ &$\mathbf{AD\downarrow}$ &$\mathbf{BLEU\uparrow}$ &$\mathbf{AD\downarrow}$ &$\mathbf{BLEU\uparrow}$ &$\mathbf{AD\downarrow}$ \\
        \midrule\midrule
        Original & 30.62 & / & 30.62 & / & 30.62 & /
        \\
        \midrule
         KD	 & 30.68$_{\text{0.06}}$	 & 0.13 & 	29.91$_{\text{-0.71}}$	 & 0.71 & 	30.46$_{\text{-0.16}}$ & 	0.17

        \cr
         EWC & 	30.66$_{\text{0.04}}$	 & 0.18 & 	30.19$_{\text{-0.43}}$	 & 0.43 & 	30.60$_{\text{-0.02}}$	 & 0.06

        \cr
        CL-NMT & 	30.85$_{\text{0.23}}$ & 	0.10	 & 25.53$_{\text{-5.09}}$	 & 5.09 & 	26.87$_{\text{-3.75}}$	 & 3.75

        \cr
         Ours & 	\textbf{31.19}$_{\text{0.57}}$	 &  \textbf{0.00}	 &   \textbf{30.61}$_{\text{-0.01}}$ & 	\textbf{0.09} & 	\textbf{31.10}$_{\text{0.48}}$ &  	\textbf{0.00}

        \cr
        \bottomrule
        \end{tabular}
        \end{threeparttable}}
        \captionof{table}{Results for different architecture models in step $1$ on German-to-English datasets, averaged over six configurations.}
        \label{tab:arc_ende}
\end{table}

\begin{table}[!t]
  \centering
  \resizebox{0.48\textwidth}{!}{
  \begin{threeparttable}
  \small\setlength\tabcolsep{3pt}
          \begin{tabular}{l|lrlrlr}
          \toprule
          \multirow{2}[3]{*}{\bf
          Method}&
          \multicolumn{2}{c}{\bf Transformer-base}&\multicolumn{2}{c}{\bf RNN}&\multicolumn{2}{c}{\bf Transformer-big} \\
          \cmidrule(lr){2-3} \cmidrule(lr){4-5} \cmidrule(lr){6-7}
          &$\mathbf{BLEU\uparrow}$ &$\mathbf{AD\downarrow}$ &$\mathbf{BLEU\uparrow}$ &$\mathbf{AD\downarrow}$ &$\mathbf{BLEU\uparrow}$ &$\mathbf{AD\downarrow}$ \\
          \midrule\midrule
          Original & 30.62 & / & 30.62 & / & 30.62 & /
          \\
          \midrule
           KD  & 20.03$_{\text{-10.59}}$	& 10.59 & 	16.17$_{\text{-14.45}}$	 & 14.45 & 	19.56$_{\text{-11.06}}$	& 11.06
  
          \cr
           EWC & 24.29$_{\text{-6.33}}$	 & 6.33 & 	23.40$_{\text{-7.22}}$	 & 7.22	 & 25.13$_{\text{-5.49}}$	 & 5.49
  
          \cr
          CL-NMT & 11.67$_{\text{-18.95}}$ & 	18.95	 & 5.09$_{\text{-25.53}}$ & 	25.53	 & 11.90$_{\text{-18.72}}$	 & 18.72
  
          \cr
           Ours &  \textbf{30.62}$_{\text{0.00}}$ &  \textbf{0.00} & \textbf{30.62}$_{\text{0.00}}$ & \textbf{0.00} & \textbf{30.62}$_{\text{0.00}}$ & \textbf{0.00}
          \cr
          \bottomrule
          \end{tabular}
          \end{threeparttable}}
          \captionof{table}{Results for malicious models in step $1$ on German-to-English datasets, averaged over six configurations.}
          \label{tab:malicious_ende}
  \end{table}

\subsection{Comparison with Multi-teacher Knowledge Distillation}
\label{app:ensemble}

\begin{figure*}[!t]
  \centering
  \small
  \begin{threeparttable}
    \begin{tabular}{l|cccccc}
    \toprule
    \textbf{Method} & \textbf{BCDE$\rightarrow$A} & \textbf{CBDE$\rightarrow$A} & \textbf{ACDE$\rightarrow$B} & \textbf{CADE$\rightarrow$B} & \textbf{ABDE$\rightarrow$C} & \textbf{ABDE$\rightarrow$C}   \\
    \midrule
     Multi-teacher KD  & 45.72 & 45.72 & 26.94 & 26.94 & 18.82 & 18.82 
    \cr
     Ours &  \textbf{45.89}  &  \textbf{46.08} &  \textbf{28.49} &  \textbf{28.51} &  \textbf{19.15} &  \textbf{18.98}
    \cr
    \bottomrule
    \end{tabular}
    \end{threeparttable}
    \captionof{table}{BLEU scores of multi-teacher KD and our method. Multi-teacher KD violates the task definition and is not applicable in our scenario.}
      \label{tab:ensemble_total}
\end{figure*}

Multi-teacher distillation differs with our method in three aspects:

\begin{itemize}
    \item \textbf{Applicable scenario.} Vanilla multi-teacher distillation averages the outputs of all teacher models as the target distribution, which requires all teacher models available at the same time, violating the task definition that a sequence of teacher models are distilled in many steps. It is impossible to get all teacher models at early steps. Thus, vanilla multi-teacher distillation cannot be used as a baseline method.
    \item \textbf{Robustness.} Vanilla multi-teacher distillation averages the output of all teacher models and is vulnerable to $D^{-}_{\mathrm{trans}}$.
    \item \textbf{Storage requirement.} The memory footprint of vanilla multi-teacher distillation will exceed the available memories of GPUs as the number of teacher models increases. A straightforward way to alleviate the problem is storing the output of teachers in a similar way as the aforementioned knowledge inheritance. However, it is non-trivial to achieve a good balance between storage requirement and performance.
    Let $|D_{\mathrm{trans}}|$ be token numbers of target sentences in $D_{\mathrm{trans}}$, $N_{\mathrm{step}}$ be the number of steps, and $N_{\mathrm{\mathcal{V}}}$ be the output vocabulary size. The following three high-potential methods all face problems: 
    \begin{itemize}
        \item \emph{Storing logits}: It leaves room for integrating knowledge filtration. However, its storage requirement is$|D_{\mathrm{trans}}|\cdot N_{\mathrm{\mathcal{V}}}\cdot N_{\mathrm{step}}$, which is impractical since $N_{\mathrm{step}}$ can be arbitrarily large.     
        
        \item \emph{Storing top-1 tokens}: It requires constant storage of $|D_{\mathrm{trans}}|$. However, knowledge filtration is hard if not impossible to be developed.
        
        \item \emph{Storing moving average of the logits}: Its storage requirement is also constant, i.e., $|D_{\mathrm{trans}}|\cdot N_{\mathrm{\mathcal{V}}}$. However, it is prone to $D^{-}_{\mathrm{trans}}$ and knowledge filtration is also hard to be integrated.
    \end{itemize}  
    In contrast, our knowledge inheritance has been shown to be effective and requires a constant size of storage ($|D_{\mathrm{trans}}|\cdot N_{\mathrm{\mathcal{V}}}$). Moreover, we still try our best to apply multi-teacher distillation on \KANMT by storing the output logits of all teacher models. The results in Table~\ref{tab:ensemble_total} show no performance advantage over our method. 
\end{itemize}

Overall, our proposed method is superior to vanilla multi-teacher distillation for \KANMT.

\section{Exploring Loss Function of Knowledge Filtration}
In this section, we will study the effects of inverse KL loss and modifying sources of knowledge filtration loss.
All experiments are conducted under the homogeneous model setting on the Chinese-to-English language pair.

\subsection{Inverse KL Loss}

Trivial KL loss is zero-avoiding and concentrates on a single mode,  while inverse KL loss covers the broad range and is zero-pursing.
\begin{equation}
    \mathrm{KL}(P_1||P_2)=-\sum_i P_1(i) \ln \frac{P_2(i)}{P_1(i)}
\end{equation}
\begin{equation}
    \mathrm{InvKL}(P_1||P_2)=-\sum_i P_2(i) \ln \frac{P_1(i)}{P_2(i)}
\end{equation}
The BLEU scores of the trivial KL loss and inverse KL loss are $31.07$ and $30.29$, respectively. Therefore, the trivial KL loss performs better in our task.

\subsection{Effect of Knowledge Filtration Loss}
In the knowledge filtration loss $\ell_{\mathrm{KF}}$, $\ell_{\mathrm{KD}}$ motivates the model to learn from student, while $\ell_{\mathrm{NEG}}$ motivates the model to learn against student. Intuitively, $\ell_{\mathrm{NEG}}$ is calculated from poor predictions from teacher models, and these error-prone distributions might provide empirical knowledge that motivates the student model not to make the same mistakes as the teachers. The output distributions from the student model could be improved by pushing them away from poor distributions of corresponding tokens output by teacher models. Conversely, random distributions are not beneficial to the student model. To verify this, we replace $\ell_{\mathrm{NEG}}$ with the following noise:
\begin{itemize}
    \item \textbf{Noise sampled from uniform distribution.}$\quad$ The probability distribution that replaces the original negative loss is obtained by sampling from a uniform distribution and passing through a softmax layer.
    \item \textbf{Noise sampled from normal distribution.}$\quad$The probability distribution that replaces the original negative loss is obtained by sampling from a normal distribution and passing through a softmax layer.
    \item \textbf{Noise from shuffled batch.} $\quad$ We randomly pick up a prediction distribution as the noise distribution from the batch from which the original negative sample is. It is worth noting that this kind of noise is usually single-peaked,  high-confidence, and more similar to the original negative sample compared to the above two noises. 
    \begin{itemize}
        \item \textbf{Attached}$\quad$The negative KD loss here is included when calculating the gradient of the sampled negative noise sample;
        \item \textbf{Detached}$\quad$The negative KD loss here is excluded when calculating the gradient of the sampled negative noise sample.
    \end{itemize}
\end{itemize}

\begin{table}[!t]
  \centering
  \small\setlength\tabcolsep{3pt}
  \begin{threeparttable}
  \label{tab:performance_comparison4}
    \begin{tabular}{lrr}
    \toprule
 {\bf Source of Noise Sample}  & \multicolumn{1}{r}{\bf Sample Size} & \multicolumn{1}{c}{\bf BLEU}
    \cr
    \midrule
    \midrule
    Uniform distribution & 1 & 30.91
    \cr
    Normal distribution & 1 & 30.86
    \cr
    Shuffled Batch (Attached) & 1 & 29.98
    \cr
    Shuffled Batch (Attached) & 5 & 29.99
    \cr
    Shuffled Batch (Detached) & 1 & 30.52
    \cr
    Shuffled Batch (Detached) & 5 & 30.62
    \cr
    \midrule
    Negative KD Loss & & \textbf{31.07} 
    \cr
    \bottomrule
    \end{tabular}
    \end{threeparttable}
      \caption{Results for replacing negative KD loss with noise sample. The metrics are averaged over six configurations. }
      \label{tab:noise}
\end{table}

As shown in Table~\ref{tab:noise}, performance is degraded when negative KD loss is derived from noise, thus illustrating that negative samples assist  student models in avoiding errors and improving the efficiency of knowledge distillation.

\end{document}